\documentclass[journal]{IEEEtran}

\ifCLASSINFOpdf
   \usepackage[pdftex]{graphicx}
  \graphicspath{{../Figures/}}
   \usepackage{subfigure}
    \usepackage{algorithm}
    \usepackage{algpseudocode}
    \usepackage{amsmath}
    \usepackage{etoolbox}
    \usepackage{caption}
    \usepackage[dvipsnames]{xcolor}
    \usepackage{mathtools}
  
\else
  
\fi

\begin{document}

\title{Torso-Based Control Interface for Standing Mobility-Assistive Devices}

\author{Yang~Chen$^{1 \dagger}$,~\IEEEmembership{Member,~IEEE,}
        {Diego}~{Paez-Granados}$^{2}$,~\IEEEmembership{Member,~IEEE,}
        {Modar}~{Hassan}$^{1}$,~\IEEEmembership{Member,~IEEE,}
        and~Kenji~Suzuki$^{1}$,~\IEEEmembership{Member,~IEEE}% <-this % stops a space

\thanks{$^\dagger$ is the corresponding author.}
\thanks{$^{1}$ Y. Chen, M. Hassan and K. Suzuki are with the Institute of Systems and Information,  University of Tsukuba, Japan. 
        {\tt\small chenyang@ai.iit.tsukuba.ac.jp, modar@iit.tsukuba.ac.jp, kenji@ieee.org}}  %  RAS PIN: 243980, 213730,}
\thanks{$^{2}${ D.} {Paez-Granados} is with Spinal Cord Injury \& Artificial Intelligence Lab (SCAI), Swiss Paraplegic Research (SPF) and ETH Zurich, Switzerland.
        {\tt\small dfpg@ieee.org}}
}

% The paper headers
\markboth{Accepted by IEEE/ASME Transactions on Mechatronics,~Vol., No., MONTH YEAR}%
{Shell \MakeLowercase{\textit{et al.}}: Bare Demo of IEEEtran.cls for IEEE Journals}

% make the title area
\maketitle

% As a general rule, do not put math, special symbols or citations
% in the abstract or keywords.
\begin{abstract}
Wheelchairs and mobility devices have transformed our bodies into cybernic systems, enhancing our well-being by enabling individuals with reduced mobility to regain freedom. Notwithstanding, current interfaces of control primarily rely on hand operation, therefore constraining the user from performing functional activities of daily living. In this work, we propose a design of a torso-based control interface with compliant coupling support for standing mobility assistive devices. 
We consider the coupling between the human and robot in the interface design. The design includes a compliant support mechanism and mapping between the body movement space and the velocity space. We present experiments including multiple conditions, with a joystick for comparison with the proposed torso control interface. The results of a path-following experiment demonstrated that users could control the device naturally using the hands-free interface, and the performance was comparable with the joystick, with $10\%$ more consumed time, an average cross error of $0.116 m$ and $4.9\%$ less average acceleration.
In an object-transferring experiment, the proposed interface demonstrated a clear advantage when users needed to manipulate objects during locomotion.
Lastly, the torso control scored $15\%$ less than the joystick on the system usability scale for the path-following task but $3.3\%$ more for the object-transferring task. 
\end{abstract}

% Note that keywords are not normally used for peerreview papers.

% For peer review papers, you can put extra information on the cover
% page as needed:
% \ifCLASSOPTIONpeerreview
% \begin{center} \bfseries EDICS Category: 3-BBND \end{center}\begin{center} \bfseries EDICS Category: 3-BBND \end{center}
% \fi
%
% For peerreview papers, this IEEEtran command inserts a page break and
% creates the second title. It will be ignored for other modes.
\IEEEpeerreviewmaketitle

\vspace{-0.3cm}
\section{Introduction}

\IEEEPARstart{I}{ndividuals} with spinal cord injuries (SCI) commonly utilize assistive devices, such as powered wheelchairs and standing mobility devices, to aid in locomotion. These mobility devices typically rely on manual operation for control. For instance, the widely used Whill wheelchair from Japan \cite{whill}, a standing wheelchair from Switzerland \cite{LEVO}, and a standing mobility device \cite{gyrolift} all employ a joystick as the primary control interface. However, the reliance on hand-operated controls limits the user’s ability to perform other tasks with their hands, in contrast to individuals without physical impairments.

Researchers have explored various approaches to alleviate the need for users to rely on their hands for mobility device operation, which can be broadly categorized into autonomous control and voluntary control. In the domain of autonomous control, studies have focused on autonomous navigation systems based on global mapping for long-distance mobility \cite{niijima2019real} or local controllers designed for specific tasks such as wheelchair docking \cite{Chen_2021}. In contrast, voluntary control involves the use of hands-free control interfaces, as outlined by Ashok \cite{ashok2017high}. While voluntary control is relatively straightforward and allows users full control over their navigation, it is currently the preferred option for most daily activities. Additionally, recent literature has highlighted increasing interest in shared control strategies, which seek to strike a balance between autonomous and voluntary control \cite{deng2019bayesian}.

This work presents a novel hands-free control interface within the category of voluntary control, designed to enable dexterous and seamless control of standing mobility devices for wheelchair users. The target end-user population for this approach includes wheelchair users with lower limb impairments but functional upper body abilities, such as individuals with spinal cord injuries (SCI) below the T12 level, as well as elderly wheelchair users. The T12 level was specifically selected because individuals with SCI above this level are more likely to experience deficits in hip muscle control, which may be necessary for the proposed control method. Building on our previous work \cite{chenTorso}, the current method introduces an improved, easily wearable coupling mechanism that provides upper body support while enabling velocity control of the wheelchair. Additionally, the sensing and control systems have been redesigned from earlier iterations to accommodate the new design and support mechanism, resulting in an expanded velocity space and more stable control.

\begin{figure}[t]
     	\centering
		\includegraphics[width=7cm]{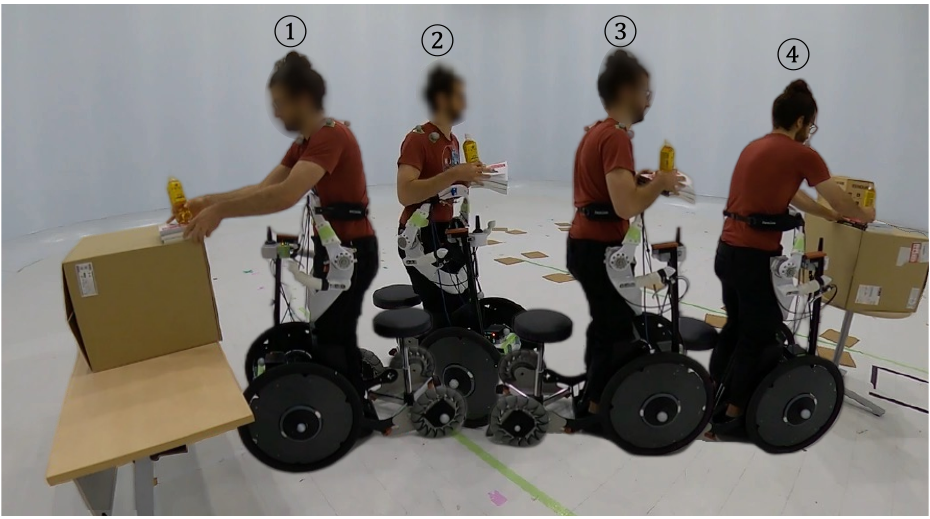}
      	\caption{Overview of the torso control, motion sequence of one participant controlling the device while transferring objects from one table to another.}
      	\label{fig:Qolo}
       \vspace{-0.4cm}
\end{figure}

\vspace{-0.3cm}
\section{Related works}
The residual upper body functions of a wheelchair user should be considered to realize hands-free control. In this section, we review the related literature and categorize the different approaches based on the anatomy of the used residual upper body function.

\vspace{-0.3cm}

\subsection{Control interfaces for robotic walkers}
Robotic walkers serving as mobility aids often feature hands-free control. For example, \cite{ji2021design} used force and torque sensors to measure pelvic motion, \cite{zhao2020smart} detailed the use of a laser scanner to measure gait patterns for human intention estimation. However, due to the severe impairment of the lower body, these methods are not suitable for most wheelchair users.

\vspace{-0.3cm}

\subsection{Neck and head-based approach}
Researchers have investigated using the neck and the head to control a wheelchair due to their multiple degrees of freedom (DOF), which has the potential to realize dexterous control.
For example, researchers have investigated using eye movement tracker \cite{dahmani2020intelligent}, brain-machine interfaces (BCI) \cite{deng2019bayesian}, head array \cite{rechy2012head}, sip-and-puff control \cite{grewal2018sip}, tooth click \cite{Simpson2008}, tongue control \cite{kong2019stand, lontis2021wheelchair}, sniff control devices \cite{plotkin2010sniffing}, voice command recognition system \cite{nishimori2007voice}, and neck EMG signal control \cite{choi2006new} as inputs for wheelchair control.
All of the above solutions can enable hands-free control, and they are preferable in cases of severe mobility impairment (cervical injury level) of the user. 
However, since the neck and head are heavily used in social life using them as control inputs for a wheelchair is not preferable for non-severe cases.
In addition, non-obtrusive interfaces such as BCI or voice control do not offer yet precise control in wheelchairs.

\vspace{-0.3cm}
\subsection{Upper-body based approach}
Researchers have explored the use of full upper body posture to control wheelchairs \cite{Chronus, UNI-ONE}. Many of these methods utilize a self-balancing approach similar to the `Segway' \cite{Nguyen2004} technology designed for able-bodied individuals. However, ensuring the safety of such self-balancing wheelchairs poses a significant challenge.
Another solution was presented in \cite{wakita2012riding}, which detected a rider's change in posture using pressure sensor sheets attached to a backrest and seat to determine the operation intention. However, an elaborate evaluation of control is yet to be presented, and standing mobility was not considered.

\subsubsection{Shoulder based approach}
Recent research suggested using a set of Inertial Measurement Units (IMU) to create a shoulder motion-based control interface for cervical-level SCI \cite{Thorp2016}. This method shows good potential for seamless control over a wheelchair. however, using the shoulder joint deteriorates the hands-free quality of the control method requiring sensors to be attached to the user's body which presents an inconvenience for daily usage. 

%%%%%%%%%%%%%%%%%%%%%%%%%%%%%%%%%%%%%%%
\subsubsection{Central torso-based approach}
In this paper we refer to the lower thoracic and abdominal segments as the central torso \cite{moore2018clinically}.
The central torso appears to be more stable than the head and neck segments during walking \cite{cromwell2001sagittal}, also, the central torso can be less interfered with by most daily activities. Therefore, it is suitable for generating control input.
One central torso-based control interface has been investigated by our group \cite{Eguchi2018}, two potentiometers were placed lateral to the user's hipbones and connected by leaf spring-like links to the torso, allowing the user's upper-body motion to serve as control inputs through the differential tilting degree. However, this interface requires the user to use both hands to support the upper body weight due to the lack of an upper body support mechanism. In a later version of our standing mobility device \cite{Paez2018}, the torso support mechanism was added. With pressure sensors attached on the inner side of the torso support mechanism, we can control the device using the detected body postures as proposed in \cite{chenTorso}. A similar method was also applied on a sitting wearable mobility device \cite{chen2023wemo}, but no quantitative evaluation was conducted.
Preliminary experiments \cite{chenTorso} have shown that most participants were able to control the device intuitively. However, as a preliminary design, the focus was on the sensing surface and algorithm, and the influence of the coupling mechanism on the control performance was not considered.
Without a suitable coupling mechanism design and control method, the contact between the human torso and the sensing surface would be unstable, resulting in the robot's velocity becoming unstable as well.

\vspace{-0.3cm}
\subsection{Proposal}
In this work, we proposed a new coupling mechanism and the corresponding control algorithm based on user-centered design. Furthermore, we designed experiments to evaluate the precision of control, usability, and comfort of the newly proposed interface for assistive mobility devices.
The contributions of this article are as follows: 
\begin{enumerate}
    \item The design of a compliant coupling mechanism with stiffness suitable for comfortable support and control. 
    \item The sensing and mapping method for providing a hands-free control.
    \item Quantitative and qualitative evaluation of the usage of hands-free control interface with a baseline of joystick control.
\end{enumerate}

\vspace{-0.3cm}
\section{Methodology}
% Explain the general methodology and the difference from the previous investigation.

\begin{figure}[]
     	\centering
		\includegraphics[width=1\linewidth]{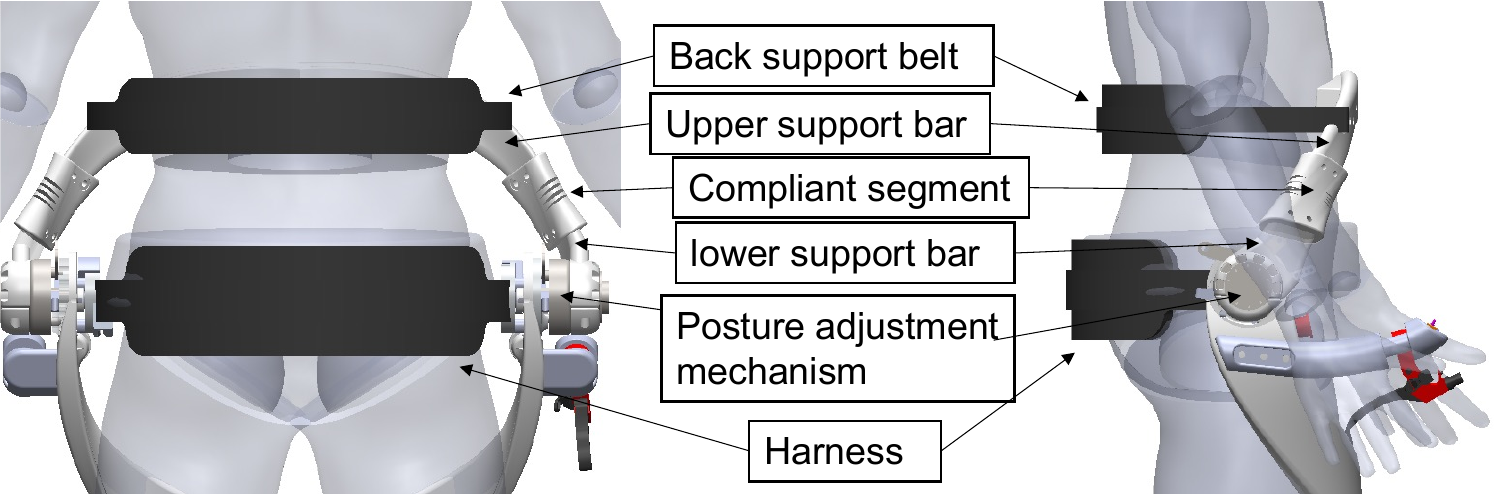}
      	\caption{The proposed interface and its components. }
      	\label{fig:mechanism}
       \vspace{-0.4cm}
\end{figure}

\subsection{Compliant coupling mechanism}\label{sec_support}
The proposed compliant coupling mechanism serves two purposes to provide a compliant support force for the upper body and to provide an interface for controlling the device.
The mechanism is composed of 6 parts as shown in Fig. \ref{fig:mechanism}:
A back support belt, an upper support bar, a compliant segment, a lower support bar, a posture adjustment mechanism, and a harness to support the user’s hip joints. 

The posture adjustment mechanism adopts a hub brake with a reversible lever (HC-CR hub brake 20H, KARASAWA, Japan). The brake is installed on the hip joint. A hand gripper is installed at a hand-releasing position. The user is able to manually adjust the angular position of the entire interface by pressing the hand gripper to reach a comfortable posture. After releasing the hand gripper, the position of the support mechanism is locked by the brake.

The compliant segment serves multiple purposes, which are detailed below:

\begin{itemize}
    \item The compliant segment generates a force increasing with the bending angle to support the upper body weight.
    \item The compliant property affords a comfortable interface compared to a solid mechanism, especially in the case of an emergency stop.
    \item When the user bends forward / backward, the upper support bar articulates around the compliant segment, which enables the use of the bending angle as a control input.
    \item The compliance affords the user two sensory feedbacks, one is the contact pressure between the torso and the interface, and another one is the bending angle which is an addition to the previous torso control system \cite{chenTorso}.
\end{itemize}

The back support belt fixes the user's back to the support bar, which ensures the movement synchronization of the torso and upper support bar. The harness is used to fix the hip joint to the standing mobility device.

\begin{figure}[]
    \begin{center}
    \includegraphics[width=1\linewidth]{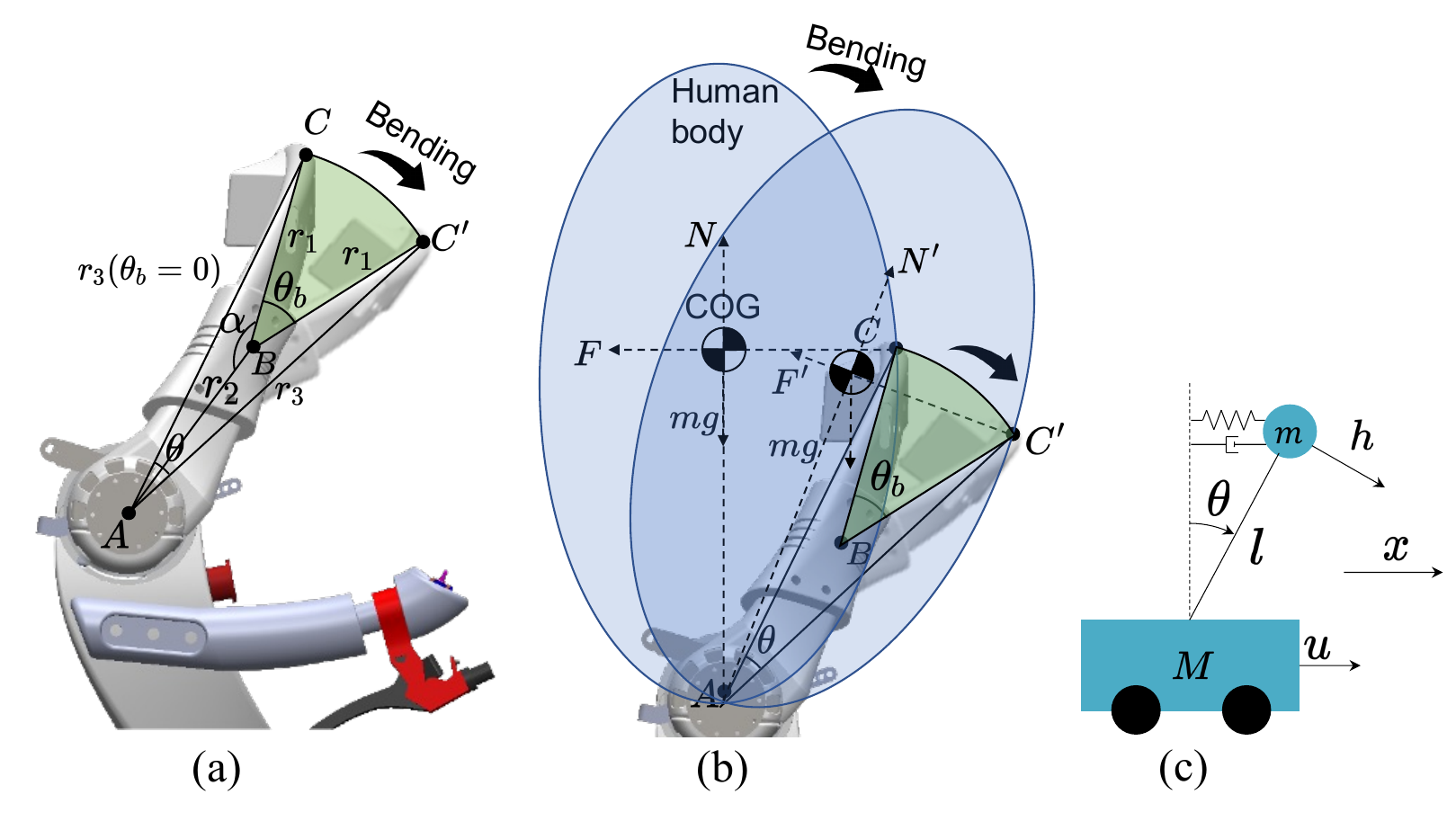}
    \caption{(a) (b) Kinematics and quasi-static force of the support mechanism. $C$ and $C^\prime$ denote the contact point between the torso and the upper support bar. $F$ and $F^\prime$ denote the supporting force from the bar, $N$ and $N^\prime$ denote the support force from the harness. (c) Simplified dynamic model of the user and mobility device in sagittal plane.}
    \label{fig:kinematics_static_f}
    \end{center}
    \vspace{-0.45cm}
\end{figure}

\subsubsection{Kinematics}\label{sec_kinematics}
The kinematics of the mechanism in the Sagittal plane are analyzed in this section. A simplified model of the mechanism is shown in Fig. \ref{fig:kinematics_static_f}(a). The length of the lower support bar $AB$ equals $r_2$. The upper support bar ($BC = r_1$) connects to the lower support bar through the compliant segment at point $B$, when there is no external force applied, the angle of $ABC$ equals $\alpha$. Owing to the compliant segment, the upper support bar rotates around $B$ point, from $BC$ to $BC^\prime$, and the rotation angle is denoted as $\theta_b$. Owing to the harness, the user's torso rotates around $A$ point, the angle is denoted as $\theta$ which is the direct sensory feedback to the user.
The relation of $\theta_b$ and $\theta$ is denoted in (\ref{equ:r3}), it is close to linear when $\theta_b$ is within (0, 25) degrees, suggesting that $\theta_b$ can be directly used as a control input. 
$r_3$ is the distance between contacting point $C^\prime$ and hip joint $A$. The length of $r_3$ changes slightly with $\theta_b$, meaning that there may exist a tangential force between the torso and support bar at the contact point $C^\prime$.

\begin{equation}
\left \{\begin{array}{l}
r_3=\sqrt{r_1^2+r_2^2-2 r_1 r_2 \cos (\alpha+\theta_b)} \\
\theta=\operatorname{arc}\left(\frac{r_1 \sin \alpha}{r_3(\theta_b=0)}\right)+\operatorname{arc}\left(-\frac{r_1 \sin (\alpha+\theta_b)}{r_3}\right)
\end{array}\right.
\label{equ:r3}
% \vspace{-0.2cm}
\end{equation}

\subsubsection{Quasi-static force analysis}\label{sec_static_force}
The quasi-static force happening in the sagittal plane is denoted in Fig. \ref{fig:kinematics_static_f}(b). $m$ denotes the upper body weight.
If we consider a 60 kg healthy participant, the upper body weight takes around 54.5\% \cite{de1996adjustments}. Combining $F_d = mg\sin\theta$ with (\ref{equ:r3}), we can have the relationship between the ideal force $F_d$ for fully supporting the upper body weight and $\theta_b$.
In theory, a stiffer compliant segment is better for supporting the upper body weight.

\subsubsection{Selection of suitable stiffness}

\begin{table}
\vspace{-0.45cm}
\centering    
\begin{tabular}{l|l|l}
\hline & Meaning & Value/unit \\
\hline$x$ & robot displacement & m \\
\hline$\dot{x}$ & robot velocity & m/s \\
\hline$\theta$ & human torso posture & rad \\
\hline$\dot{\theta}$ & angular velocity & rad/s \\
% \hline & Meaning & Value/unit \\
\hline$t$ & time & s \\
\hline$m$ & mass of the pendulum & 32.7kg \\
\hline$M$ & mass of cart & 75kg \\
\hline$l$ & length of pendulum & 0.25m \\
\hline$g$ & gravity coefficient & 10$m/s^2$ \\
\hline$h$ & human applied force & $N$\\
\hline$h_m$ & maximum human applied force & $33.65 N$\\
\hline$F_c$ & contact force & N \\
\hline$k_s$ & spring coefficient & $2000\sim3000$ N/s \\
\hline$k_c$ & damper coefficient & 30 $N\cdot s/m$\\
\hline$k_d$ & friction coefficient & 1 \\
\hline $k_2$ & control input coefficient & 0.02 \\
\hline $k_3$ & control input coefficient, real & \\
 & intention & 100 \\
\hline
\end{tabular}
\caption{Parameter for dynamics modeling.}
\label{table:dynamics}
\vspace{-0.45cm}
\end{table}

Selecting a suitable stiffness for the complaint mechanism is important to ensure stable control of the mobility device under dynamic settings.
A simplified dynamics model of the mobility device and compliant support mechanism is shown in Fig. \ref{fig:kinematics_static_f}(c). The inverted pendulum represents the weight of the upper body while the lower body and the robot are represented by a cart. The spring and damper are used to represent the coupling relationship between the torso and the upper body support system. $h$ is modeled to represent the external force generated by the human's torso. The system states and the parameters of the dynamics of the coupling system are shown in TABLE \ref{table:dynamics}. The dynamics derivation is shown in Appendix \ref{Appendix}.

In order to fabricate a compliant segment with suitable characteristics we simulated under different spring coefficients. In addition to the dynamics of the human-robot coupling system, the high-level and low-level controllers are both modeled as proportional ones as shown in (\ref{equ:high-low-level}). 
The human intention is expressed by applying an external force $h\left(h<h_m\right)$ on the pendulum, while the estimated intended velocity (denoted as $v_{ref}$) by the robot is through the contact force $F_c$ between the torso and the spring. The driving force is denoted as $u$.
\begin{equation}
v_{ref} = k_2 \cdot F_c, \: 
u = k_3 \cdot (v_{ref}-\dot{x})
\label{equ:high-low-level}
\end{equation}

\begin{figure}[!t]
      \centering
  		\subfigure[$\theta - t$]{\includegraphics[width=0.48\linewidth]{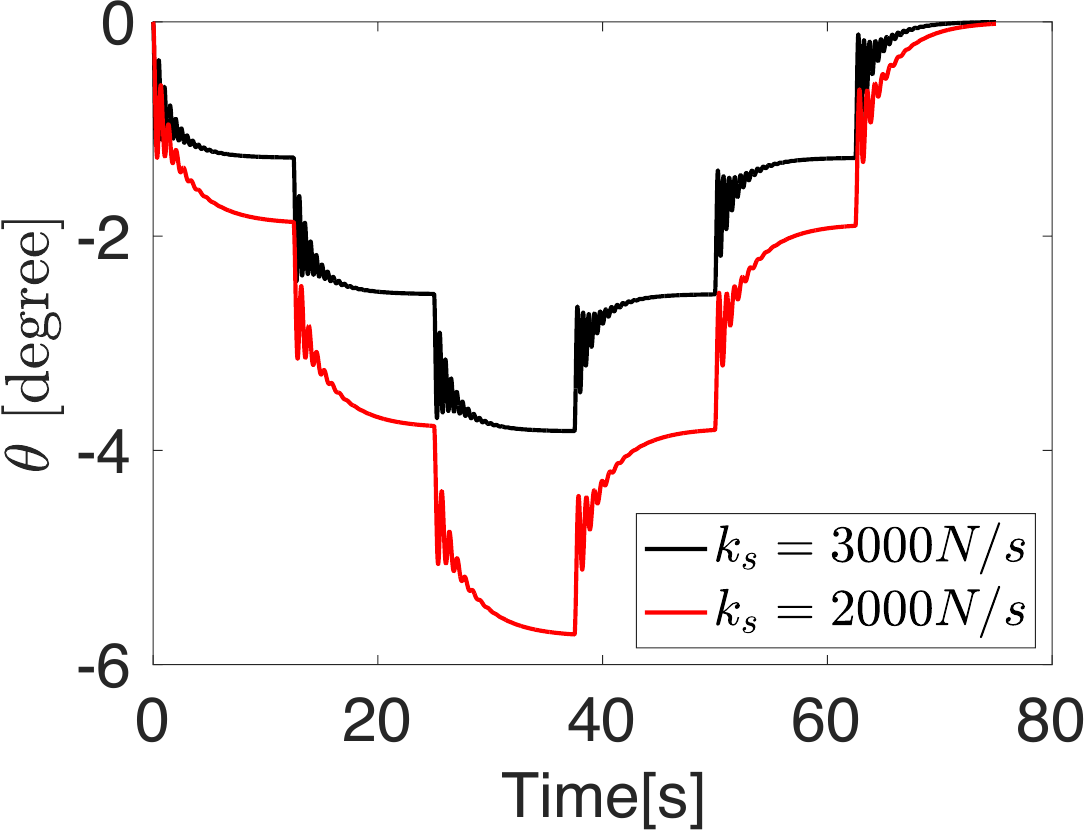}
		\label{fig:theta_800_2000}}
		\hfil
		\subfigure[$\Dot{x} - t$]{\includegraphics[width=0.48\linewidth]{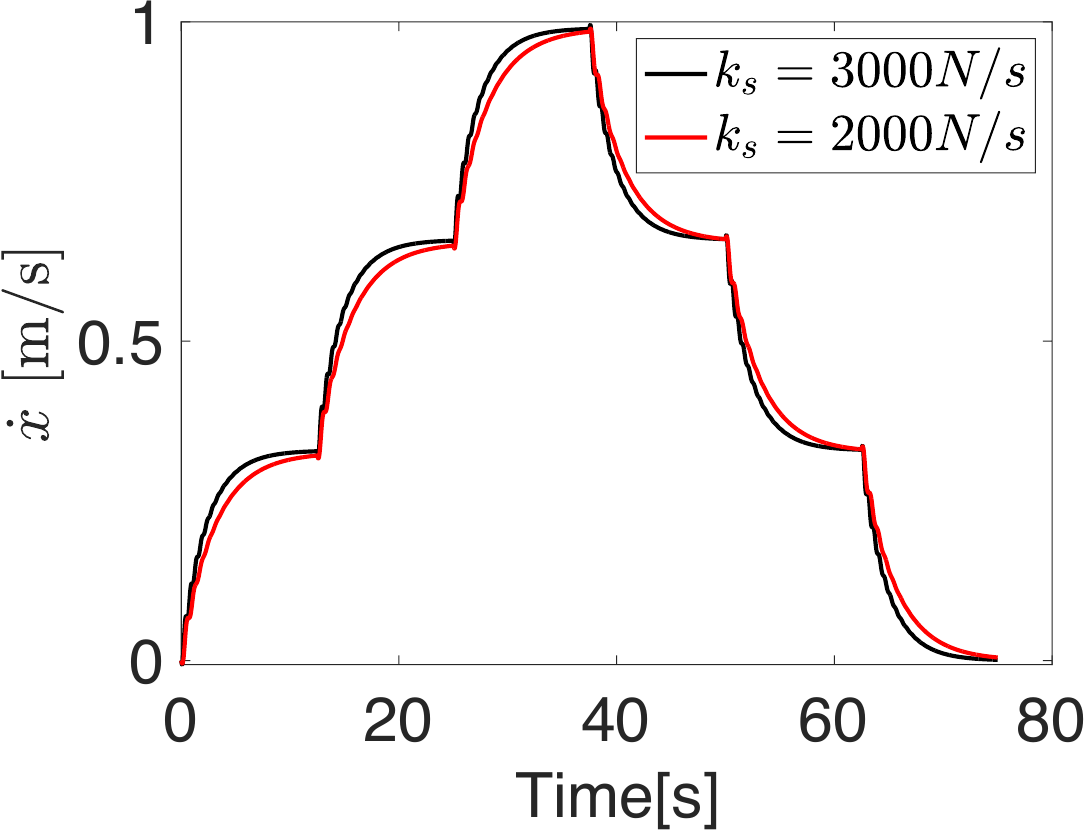}%
		\label{fig:xdot_800_2000}}
      \caption{Changed in $\theta$ and $\Dot{x}$ with time for spring coefficient $k_s$ of 2000 $N/s$ and 3000 $N/s$ respectively. }
      \label{spring:800_2000}
      \vspace{-0.4cm}
   \end{figure}

From the simulation, we found that the minimum spring coefficient for fully supporting upper body weight (32.7kg) is $1308N/s$.
An example of the human operation process is set up as follows. The human applies force from 0 to maximum then from maximum to 0 divided by six stages, staying at each stage for 10 seconds. The result of $\theta$ and $\Dot{x}$ when the spring coefficient $k_s$ equals to 2000 $N/s$ and 3000 $N/s$ are depicted in Fig. \ref{fig:theta_800_2000} and Fig. \ref{fig:xdot_800_2000} respectively. We calculate the average angular acceleration ($A_{aa}$) of the torso to indicate the degree of vibration using (\ref{equ:aaa}).
\begin{equation}
A_{aa}=(1 / N) \sum_{i=1}^N\left\|\ddot{\theta}\right\|
\label{equ:aaa}
\end{equation}

For the case of the spring coefficient equaling 2000 N/s and 3000 N/s, $A_{aa}$ equals 4.9644 $rad/{s^2}$ and 9.3132 $rad/{s^2}$ separately, which corresponds to the finding from Fig. \ref{spring:800_2000}: the compliant segment with lower stiffness causes less vibration than the one with higher stiffness. However, a noticeably higher delay is also observed with the more compliant one. Since a higher vibration would cause discomfort to the user, and a delay within a certain range is considered to be acceptable, we decided to use the more compliant segment in the developed prototype. The final implemented stiffness of the compliant segment was tuned considering the user's comfort through preliminary driving tests on the device.

%%%%%%%%%%%%%%%%%%%%%%%%%%%%%%%%%%%%%%%%%

\vspace{-0.3cm}
\subsection{Control Method: Sensing and Mapping}\label{sec_sm}
In this work, we separate the velocity magnitude control (from the IMUs) and the velocity direction control (from the FSRs). 
We then integrated both controls with a two-layer mapping from sensor space to robot motion space.
This results in a stable control and is a unique approach to this research, to the best of our knowledge.

The sensing system consists of an array of Force Sensitive Resistors (FSRs) and two Inertial Measurement Units (IMUs). The FSR array is used for pressure sensing of the human body and is designed to cover the waistline of an adult. The spatial distribution of the sensors is similar to our previous work \cite{chenTorso}. Each FSR is $4 \times 4\mathrm{~cm}$, with approximately 4 mm spacing between them, covering a total of around 21.6 cm. A support layer provides a solid surface for the FSRs, with a curvature designed initially based on a 3D adult human body model in Solidworks and then adjusted according to real tests to ensure proper contact with the torso. One of the IMUs is attached to the exterior surface of the upper support bar, while the other is attached to the base of the robot. The relative bending angle between the two IMUs is used as the bending angle.

The space mapping relation in this work is as follows: user muscle space $\rightarrow$  user motion space ($\theta, \phi, \beta$) $\rightarrow$ sensor space ($\theta_b, \lambda_i$) $\rightarrow$ pointer space ($P, \delta$) $\rightarrow$ robot motion space ($v, w$). ($\theta, \phi, \beta$) represent the user's motion in sagittal, transverse, and frontal respectively. We designed the mapping from sensor space to robot motion space. The mapping from user muscle space to sensor space relies on the user to learn this by themselves through a feedback process.
With this design, we aim to minimize the user's effort and maximize the intuitiveness for control.

\begin{figure}[t]
    \begin{center}
    \includegraphics[width=0.9\linewidth]{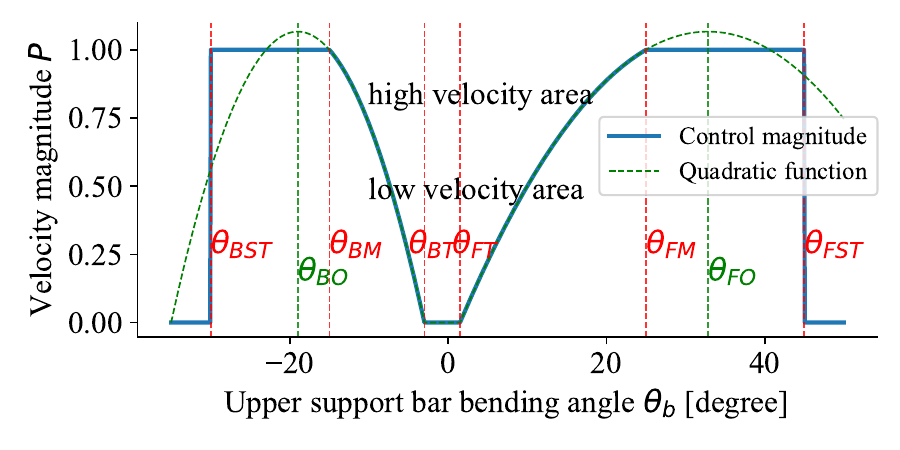}
    \caption{Mapping between bending angle $\theta_b$ and velocity magnitude $P$.}
    \label{theta_b_P}
    \end{center}
    \vspace{-0.4cm}
\end{figure}

\renewcommand{\arraystretch}{1.0} % increase the line space
\begin{table}
\centering
\begin{tabular}{l|l} 
\hline $\theta_{FT}$ & forward threshold \\
\hline $\theta_{FM}$ & forward maximum \\
\hline $\theta_{FST}$ & forward safety threshold \\
\hline $\theta_{BT}$ & backward threshold \\
\hline $\theta_{BM}$ & backward maximum \\
\hline $\theta_{BST}$ & backward safety threshold \\
\hline
\end{tabular}
\caption{Constant values in Fig. \ref{theta_b_P}.}
\label{threshold}
\vspace{-0.4cm}
\end{table}

\subsubsection{Mapping between sensor space and pointer space}
\begin{figure}[t]
    \begin{center}
    \includegraphics[width=1\linewidth]{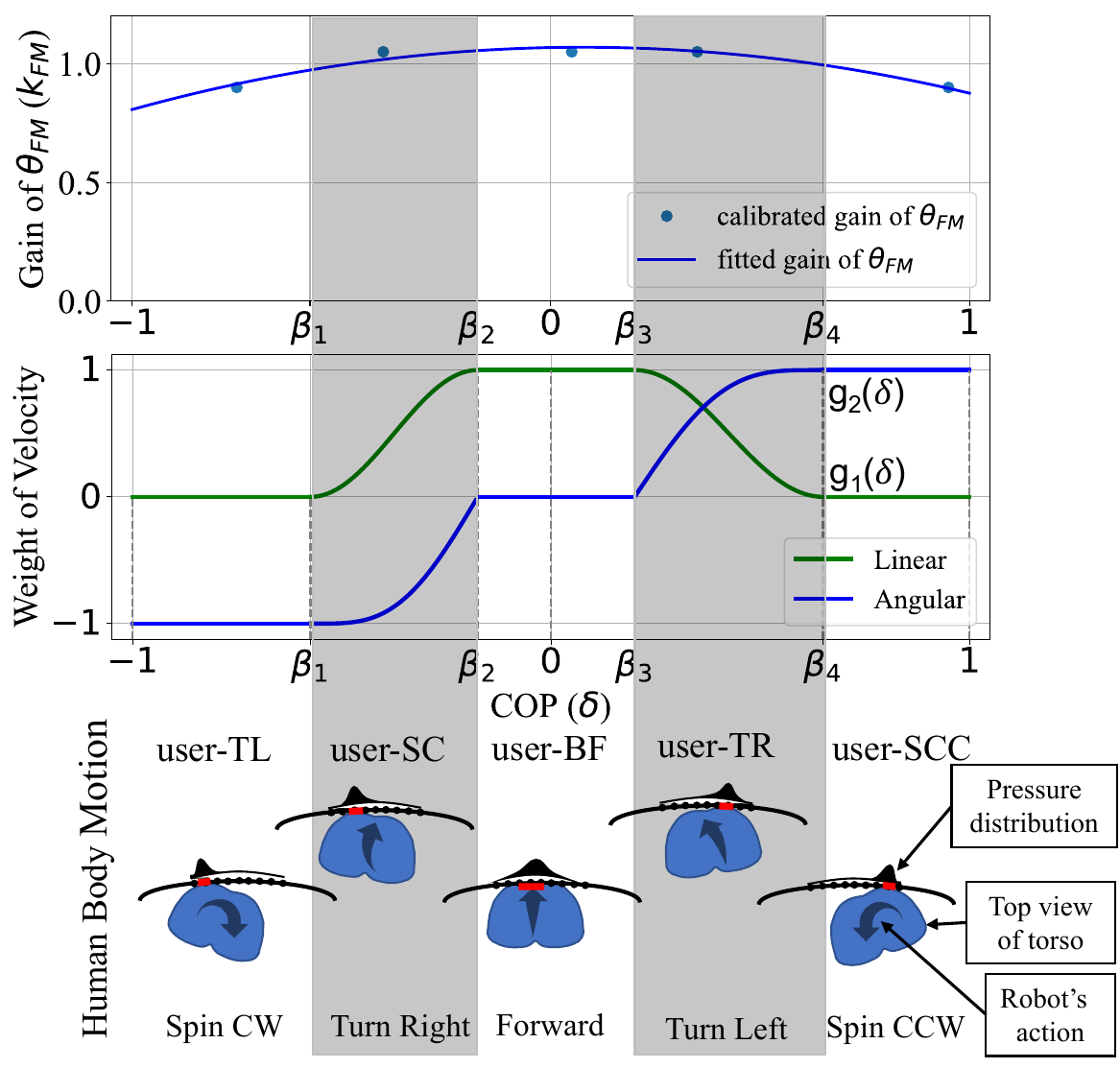}
    \caption{Upper panel: Gain of $\theta_{FM}$ ($k_{FM}$) vs COP $\delta$; middle panel: Expected weight of velocity vs COP; lower panel: Expected human posture and robot action.}
    \label{fig:fm_direction}
    \end{center}
    \vspace{-0.3cm}
\end{figure}

First, we convert the FSR reading to conductance value $\lambda$, since the relation between conductance and force is close to linear. The algorithm for calculating the center of pressure (COP) is explained in (\ref{equ:Lambda}), where the $\lambda_i$ represents the individual sensor's value, $\alpha_i$ represents the corresponding weight for each sensor that can be calibrated, and $s_i$ is the normalized sensor location. Depending on the value of $\delta$. The user's posture can be classified into 5 main categories (Fig. \ref{fig:fm_direction}): user bends forward (user-BF), user turns left (user-TL), user turns right (user-TR), user spins counter-clockwise (user-SCC), and user spin clockwise (user-SC).
\begin{equation}
\delta=\frac{\sum_{i=1}^n \alpha_i \lambda_i s_i}{\sum_{i=1}^n \lambda_i}
\label{equ:Lambda}
\end{equation}

The mapping function between the bending angle $\theta_b$ and the velocity magnitude $P\in[0,1]$ is shown in Fig. \ref{theta_b_P}, and their corresponding relationship is shown in (\ref{equ:imum}), (\ref{equ:nonlinear_m}).

\begin{equation}
P = \left\{
\begin{aligned}
&F_B(\theta_b), && \theta_{BM} < \theta_b < \theta_{BT}  \\
&F_F(\theta_b), && \theta_{FT} < \theta_b < \theta_{FM}
\end{aligned}
\right.
\label{equ:imum}
\vspace{-0.2cm}
\end{equation}

\begin{equation}
\left\{
\begin{aligned}
&F_i(\theta_b) = k_{iamp}\left[-k_{in}(\theta_b-\theta_{iO})^2 + 1 \right] \\
&\theta_{iO}=p_{im} + \theta_{iT}\\
&p_{im} = \frac{\theta_{iM}-\theta_{iT}}{k_s }\\
&k_{in} = (\frac{1}{p_{im}})^2 \\
&k_{iamp} = \frac{1}{k_s(2-k_s)}
\end{aligned}
\right., i = F, B, 
\label{equ:nonlinear_m}
\end{equation}

where $\theta_{BST}$, $\theta_{BM}$, $\theta_{BT}$, $\theta_{FT}$, $\theta_{FM}$, $\theta_{FST}$ are explained in Table. \ref{threshold}, $k_s$ (equals $0.75$ here) is the value for adjusting the ratio of the intermediate nonlinear part taken in the half branch of the parabola, in order to have a suitable gradient. $\theta_{iO}$ represents the centerline of the parabola. The intermediate nonlinear part was designed because we found that the angle detection from the IMUs is more stable in the low-speed region than in the high-speed region.
$\theta_{FM}$ and $\theta_{BM}$ are the maximum forward and backward bending angle.
A safety threshold $\theta_{BST}$ and $\theta_{FST}$ is set to avoid an accidental falling down situation. When the user bends to this angle, the robot will stop moving.

$\theta_{FM}$ should depend on the postures of the user considering that it is easier to achieve a big angle in user-BF than other postures. Therefore, in (\ref{equ:fm}), the gain $k_{FM}$ is set as a value changing with the user's posture indicated by $\delta$. $\theta_{FM}^d$ is the default value set as 25 degrees. $f_1(\delta)$ is a second-order function fitted by 5 discrete pre-defined points of $(\delta_i, k_{FM}^i)$ which can be obtained in the calibration process, as shown on the upper panel of Fig. \ref{fig:fm_direction}.

\begin{equation}
\theta_{FM} =k_{FM}\theta_{FM}^d, \: 
k_{FM} = f_1(\delta)
\label{equ:fm}
\end{equation}

\subsubsection{Mapping between pointer space and robot motion space}
For driving forward (when $\theta_b \geq 0 $), the directional control relies on the mapping from the COP to the weight distribution of velocity ($W_v\in[0,1]$ and $W_w\in[-1,1]$, middle panel of Fig. \ref{fig:fm_direction}), which is optimized from \cite{chenTorso} to increase the velocity space, resulting in the action of the robot as shown on the lower panel of Fig. \ref{fig:fm_direction}. 
The design principle is: when the user intends to control the device to turn to one side, the trunk rolls to the same side while touching the opposite side of the sensing surface, which agrees with the ﬁnding introduced in \cite{courtine2003human} that the trunk rolls toward the inner side of the turning during a human’s natural walking.
For driving backward (when $\theta_b < 0 $), the current implementation only considers one direction, and the velocity is regulated by a weight of $W_v^b$ (equals $-0.8$) for safety consideration. The overall function is shown in (\ref{equ:vw}), (\ref{equ:g1}), (\ref{equ:g2}).
\begin{equation}
\left\{
\begin{aligned}
&v = v_m \cdot P \cdot W_v(\delta,\theta_b)\\
&w = w_m \cdot P \cdot W_w(\delta,\theta_b) \\
\end{aligned}
\right.,
\label{equ:vw}
\end{equation}

\begin{equation}
W_v(\delta,\theta_b) = \left\{
\begin{aligned}
& W_v(\delta), && \theta_b \geq 0 \\
& W_v^b, &&  \theta_b < 0
\end{aligned}
\right.,
\label{equ:g1}
\end{equation}

\begin{equation}
W_w(\delta,\theta_b) = \left\{
\begin{aligned}
&W_w(\delta), && \theta_b \geq 0 \\
&0, && \theta_b < 0
\end{aligned}
\right.,
\label{equ:g2}
\end{equation}
where $v_m$ and $w_m$ are maximum linear and angular velocities. The detailed functions of weight shown in Fig. \ref{fig:fm_direction}  are denoted in (\ref{equ:linear2}), (\ref{equ:angular2}), and (\ref{equ:h1_h2}).

\begin{equation}
W_v(\delta)=\left\{
\begin{aligned}
&\frac{1}{2} + h_1(\beta_1, \beta_2) , &&  \beta_1 \leq \delta < \beta_2 \\
&\frac{1}{2} - h_1(\beta_3, \beta_4), && \beta_3 \leq \delta < \beta_4 \\
\end{aligned}
\right.
\label{equ:linear2}
\end{equation}

\begin{equation}
W_w(\delta)=\left\{
\begin{aligned}
&\frac{-1}{2} - h_2(\beta_1, \beta_2) ,&&  \beta_1 \leq \delta < \beta_2 \\
&\frac{1}{2} + h_2(\beta_3, \beta_4) ,  && \beta_3 \leq \delta < \beta_4 \\
\end{aligned}
\right.
\label{equ:angular2}
\end{equation}

\begin{equation}
\begin{aligned}
h_1(\beta_i, \beta_j) = \frac{1}{2}\sin(\frac{\pi}{\beta_i-\beta_j}(\delta-\beta_j) + \frac{\pi}{2})\\
h_2(\beta_i, \beta_j) = \frac{1}{2}\sin(\frac{\pi}{\beta_i-\beta_j}(\delta-\beta_i) + \frac{\pi}{2})\\
\end{aligned},
\label{equ:h1_h2}
\end{equation}

\begin{figure}[t]
    \begin{center}
    \includegraphics[width=1\linewidth]{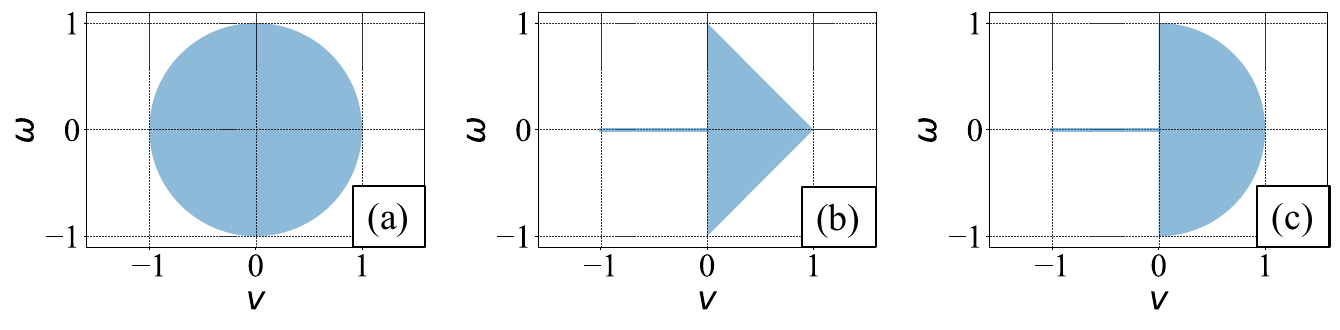}
    \caption{Normalized robot velocity space comparison, $v$: linear velocity, $\omega$: angular velocity. (a) Joystick controller. (b) Previous torso controller. (c) Current torso controller.}
    \label{fig:velocity_space_comparison}
    \end{center}
    \vspace{-0.6cm}
\end{figure}

For the final velocity command, a derivative component is added to smooth the command as shown in (\ref{equ:vwpd}), both $k_v^d$ and $k_w^d$ are set as $-0.9$.
\begin{equation}
v^\prime = v + k_v^d \dot{v}\Delta t, \:\: w^\prime = w + k_w^d \dot{w}\Delta t,\\
\label{equ:vwpd}
\end{equation}
In total, with the presented space mapping we expect to achieve an intuitive mapping from the user muscle space to the robot motion space.

The robot velocity space of the joystick controller, previous torso controller (presented in \cite{chenTorso}), and the current torso controller are shown in Fig. \ref{fig:velocity_space_comparison}. With the current design, for forward control, the linear and angular velocity space of torso control is the same as the joystick.
For backward control, the robot motion space is smaller than the joystick space.

 \vspace{-0.3cm}
\subsection{Personalization considerations}\label{sec_pc}
The proposed control method contains personalization considerations including hardware and software calibration.

\subsubsection{Hardware calibration}
For the hardware, we designed a sensor support layer to adapt to users’ body shapes which can be replaced to fit different body shapes.
For users with different body height and weight, the compliant segment can also be replaced.
The user can select a preferred posture by adjusting the posture adjustment mechanism.

\subsubsection{Software calibration}
Different users have preferred neutral posture, maximum bending angle, and postures for direction control. These can be calibrated through an off-line process which can be triggered by an onboard sensor.

For the calibration under the neutral or relaxed posture of the user, this is triggered by the user pressing the onboard sensor for 0.5s, then the program will collect 2-second reading values of each FSR sensor as well as IMU and get the mean of each sensor as zero offsets.

For the calibration of $\alpha_i$, $k_{FM}^i$, $\beta_i$, this can be triggered by double clicking the onboard sensor within 0.5s. Then the user is asked to switch from user-TR to user-TL posture with the intention to drive at the highest speed, then move in the reverse order for a set of ten transitions staying at each posture for 3 seconds (indicated by an LED on board). The data of the $i$th FSR and IMU are collected as $\overline {\Lambda_i}$ and $\Theta_{FM}$ respectively.
In total $N_s$ sets of readings are recorded, the top ten percent readings of each single sensor will be summed up and get the average value: $\overline{\lambda_{1}}, \overline{\lambda_{2}}..., \overline{\lambda_{n}}$, then based on (\ref{alpha_cal}), we could get the value of $\alpha_1, \alpha_2, ..., \alpha_n$.
After obtaining $\alpha_i$, all FSR values multiply $\alpha_i$ as the new readings. Then we calculate the COP \(\delta\) for each posture as \(\delta_1\), \(\delta_2\), \(\delta_3\), \(\delta_4\), \(\delta_5\).
The calibrated classification points are obtained as: \(\beta_1 = (\delta_1 + \delta_2)/2, \beta_2 = (\delta_2 + \delta_3)/2, \beta_3 = (\delta_3 + \delta_4)/2, \beta_4 = (\delta_4 + \delta_5)/2\).
Under each posture, the maximum bending angle $\theta_{FM}^{i}$ is calculated by averaging the values, then (\ref{kfm_cal}) is applied to obtain $k_{FM}^i$.

\begin{equation}
P_{max} = \alpha_i\overline{\lambda_i}, i = 1...n
\label{alpha_cal}
\vspace{-0.4cm}
\end{equation}

\begin{equation}
\theta_{FM}^{d} = k_{FM}^i\theta_{FM}^{i}, i = 1...5
\label{kfm_cal}
\end{equation}
%%%%%%%%%%%%%%%%%%%%%%%%%%%%%%%%%%%

\begin{figure*}[!t]
     	\centering
		\includegraphics[width=0.95\linewidth]{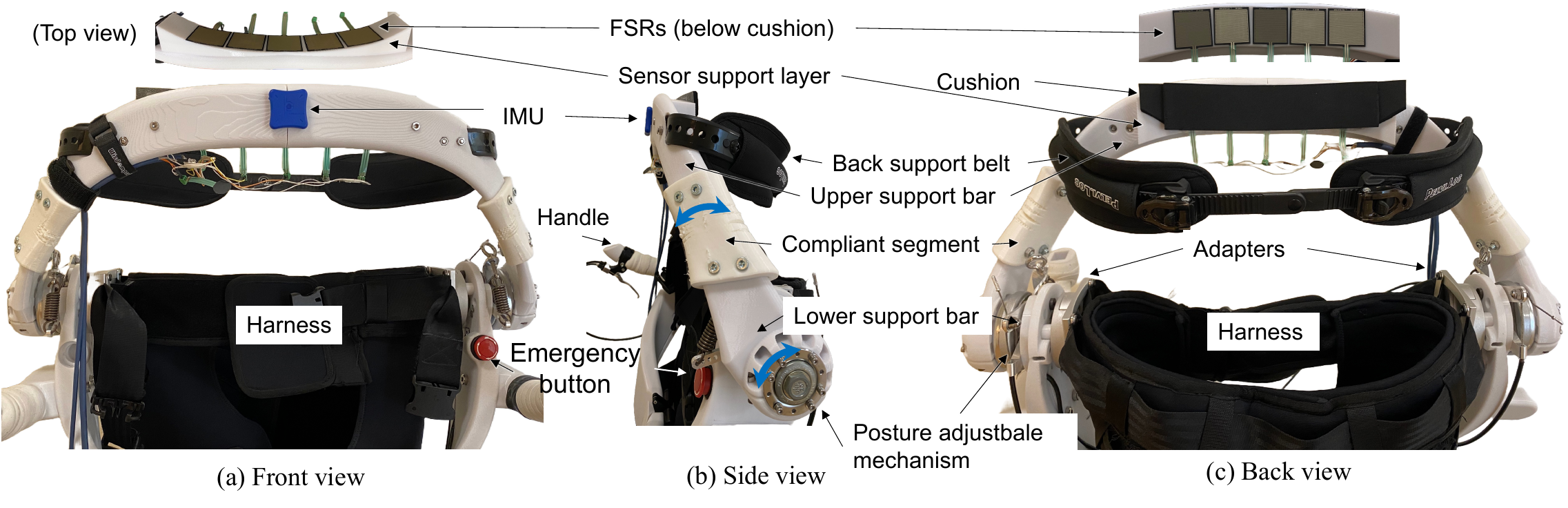}
      	\caption{The proposed interface, (a) front view, (b) side view, (c) back view. }
      	\label{fig:hardware}
       \vspace{-0.3cm}
\end{figure*}

 \vspace{-0.3cm}
\section{System Overview}\label{sec_overview}
The platform we used in this work is a modified version of Qolo, which is a hybrid wheelchair/standing mobility vehicle \cite{eguchi_thesis}. The original device does not have a torso support mechanism or a seat, therefore we modified it by adding the torso support mechanism to the hip joints and a chair to the base to realize standing and sitting navigation in this study.  
The developed torso support mechanism is shown in Fig. \ref{fig:hardware}, and the implemented control system is shown in Fig. \ref{system_overview}.

\subsubsection{Compliant torso support mechanism}
The upper and lower support bars as well as the compliant segment were manufactured with 3D printing. The support bars were fabricated with PolyCarbonate, while the compliant segment was fabricated using flexible filament, Thermoplastic PolyUrethane (TPU).
A harness (Harness Hybrid HV-120, MORITOH Co., Ltd. Japan) is used to attach the user to the hip joints with custom-made adapters and can adapt to a certain range of hip dimensions. This harness is necessary to ensure a stable connection between the user and the device while in standing posture. According to a physiatrist experienced in working with wheelchair users and individuals with SCI, the design of the device and the harness can be enhanced to better accommodate users with varying body dimensions in future developments. 

\subsubsection{Sensing system}\label{sec_sensing}
The sensor support layer is 3D printed using PolyCarbonate. The inner side of this layer is curved to resemble the human body shape for better contact. This layer can be replaced for adapting different users' waist curves. The FSRs are attached directly to the inner surface of this layer. 
A soft cushion is used to cover all FSRs. Fig. \ref{fig:hardware} shows a view of the sensing surface.

\subsubsection{Embedded control system}
\begin{figure}[t]
    \begin{center}
        \includegraphics[width=0.9\linewidth]{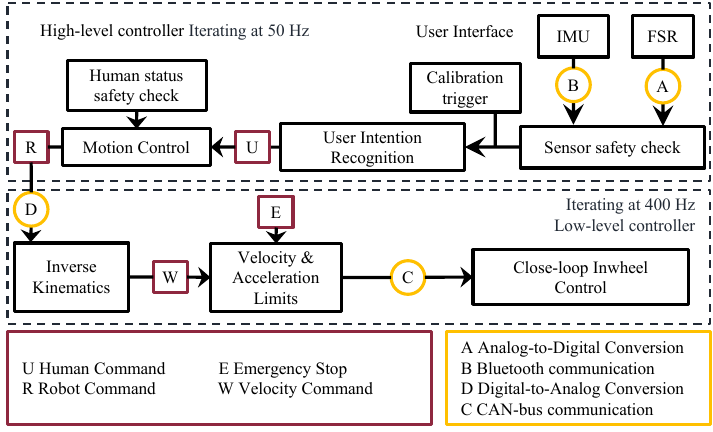}
        \caption{Control system architecture.  }
        \label{system_overview}
    \end{center}
    \vspace{-0.4cm}
\end{figure}

A Jetson AGX Xavier (Nvidia, USA) was used as a single-board computer, with which the IMUs (LPMS-B2, LP-RESEARCH Inc.) communicate through Bluetooth. A High-Precision AD/DA Board (Waveshare) was used for reading the FSRs (FSR 408, Interlink Electronics, Inc.) input and communicating through analog channels to the low-level controller. The low-level controller and the actuator are from a commercial wheelchair JWX1 (YAMAHA Motor Co., Ltd.). The overall control architecture is presented in Fig. \ref{system_overview}. The system was implemented within the architecture of the Robot Operation System (ROS).

\section{Experiments}\label{sec_experiment}

\subsection{Torso support compliance test}
A test was conducted to measure the mechanical characteristics of the torso bar compliance. The supporting force provided by the compliant segment is measured by a force gauge (ZP-2500N, IMADA Co., Ltd., Toyohashi, Japan). A towing belt was attached to the middle of the upper support bar to connect the force gauge through a steel hook. An increasing pulling force was applied to the force gauge slowly, resulting in the upper part bending. In the pulling process, the force gauge was kept perpendicular to the bar. The bending angle was measured by the IMU. This process was repeated in both forward and backward directions.

      \begin{figure}[!t]
      \centering
        \includegraphics[width=1.0\linewidth]{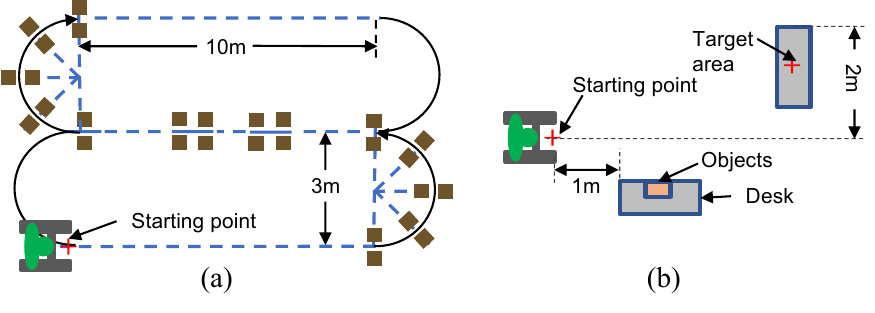}
      \caption{Top-down illustration of the experimental setup. (a) Path-following experiment. (b) Object-transferring experiment.}
      \label{exp:setup} 
      \vspace{-0.4cm}
   \end{figure} 

\subsection{User study}

A user study was conducted to verify the usability of the proposed system.
10 healthy participants (8 males, 2 females, 20s) were recruited. The inclusion criteria were: age: 20 $\sim$ 40, weight: 50 kg $\sim$ 90 kg, height: 150cm $\sim$ 190cm, no previous or current neurological or musculoskeletal disorders affecting the ability to stand up or sit down or walk, no condition that affects the body joints such as joint pain, no conditions that affect balance, ability to understand and follow the instructions of the experiments. The participants were asked to complete two tasks simulating different real-life scenarios, the first one aimed to evaluate the navigation ability, and the second one aimed to evaluate the manipulation ability while using two interfaces: the developed torso-based control and a traditional joystick.
The experiment was approved by the ethical review of the University of Tsukuba (Approval number: 2022R708-3). Before the experiments, the procedure was explained to each participant. 
For both tasks, the participants had to use the torso control interface and a joystick separately. The order of interfaces was randomized. 
For torso control, the participants were asked to relax their leg muscles and let their weight be supported by the harness while using the device, then the calibration procedure was performed. Although the sensor support layer and compliant segment can be replaced, in this study all users used the same components. The details of both tasks are shown below:
\begin{itemize}
\item Task 1: the participants controlled the device to follow an 8 shape circuit as shown in Fig. \ref{exp:setup}(a). The circuit is combined by three straight lines and 4 half-circuits. The path is defined by attaching green tape on the floor, square tiles with a height of 0.5cm and length of 30cm were attached along a part of the path as obstacles to simulate the outdoor environment. For both interfaces, the users completed two trails at a constant slow speed first, then as fast as they could. A break was given after using each interface. Before the trial, practice time was given to each participant until they felt confident to perform the task. 
\item Task 2: the participants had to transfer particular objects from one table to another one while controlling the device as shown in Fig. \ref{exp:setup}(b). The objects include a 5kg box, 3 books, and a bottle of water. The participants were free to choose any method to transfer the objects, they could also give up particular objects if they felt it was too difficult or dangerous to accomplish.
\end{itemize}

\subsubsection{Data recording}
For task 1, a motion capture system (Optitrack, NaturalPoint, Inc.) was used to capture the trajectory of the device. The in-wheel odometry is also used for recording the velocity of the robot.
For both tasks, the videos were recorded. A system usability scale (SUS) questionnaire and 6 experiment-specific questions (5-point Likert scale) were distributed to participants after each task using either interface.
The experiment-specific questions were designed with reference to standard questionnaires such as \cite{lewis1995ibm}, and each question has the same weight: (1/2/3) I felt the interface was natural/enjoyable/safe to use. (4) I felt uncomfortable using the interface. (5) I felt the interface was very demanding. (6) I felt the interface was very useful/needed.

After the participants filled out all the questionnaires, we also conducted a short interview to ask about their impressions of the two interfaces and any other feedback they wished.

\subsubsection{Performance metrics}
The performance difference between the joystick and torso control was assessed by measuring task completion time ($CT$), average acceleration ($A_{a}$), and average cross error ($A_e$).
Task completion time ($CT$) is the time used to follow the path with the highest speed. The $CT$ is calculated from the starting point until the end.
We evaluate the velocity control stability in slow speed mode from the average acceleration calculated by  (\ref{equ:aa}).
\begin{equation}
A_a=1 /(2 T)\left[\sum_{i=1}^N\left\|\dot{v}_i\right\|+\sum_{i=1}^N\left\|\dot{w}_i\right\|\right],
\label{equ:aa}
\end{equation}
 where $\dot{v_i}$ and $\dot{w_i}$ indicate linear and angular acceleration individually, $T$ is total time length, $N$ is the total sample size.
The path-following accuracy is evaluated from the cross error calculated by the average distance between the real path and the ground truth path using (\ref{equ:ae}).
\begin{equation}
A_e=1 /N\sum_{i=1}^N\left\|e_i\right\|,
    \label{equ:ae}
\end{equation}
where $e$ is defined as (\ref{equ:error}), $L: a x+b y+c = 0$ is a straight line connecting two nearest points to the location of the device $(x_c, y_c)$. $a$, $b$ and $c$ are parameters defining the straight line.
\begin{equation}
e=\frac{\left|a x_c+b y_c+c\right|}{\sqrt{a^2+b^2}}
\label{equ:error}
\end{equation}

A Wilcoxon signed rank test was applied to all performance metrics as well as the SUS to determine significant differences ($p < 0.05$) between the two interfaces.

For the experiment-specific questionnaire, we evaluate the result by counting each individual's preference instead of calculating the average score, if the participant gives a higher score to one interface than the other, then the score of this interface adds one up, if the participant gives the same score to both interfaces, then both interfaces add one up.

\vspace{-0.3cm}
%%%%%%%%%%%%%%%%%%%%%%%%%%%%%%%%%%
\section{Experimental Results and Discussions}

\begin{figure}[t]
     	\centering
		\includegraphics[width=0.8\linewidth]{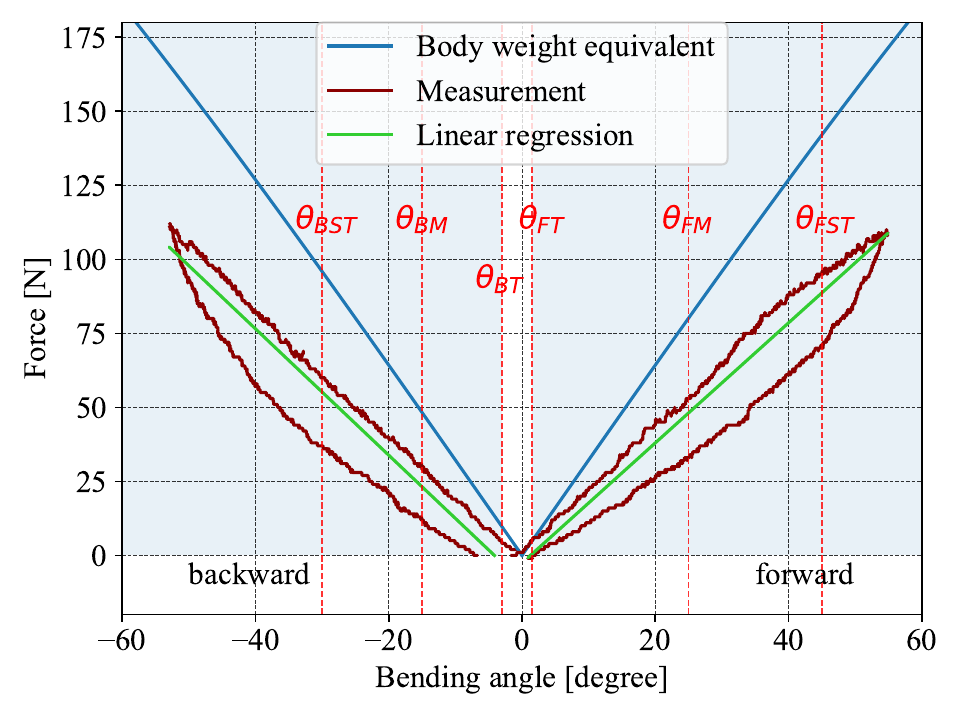}
      	\caption{The relationship between measured force and bending angle of the compliant segment.}
      	\label{fig:F-angle}
       \vspace{-0.4cm}
\end{figure}

\subsection{Torso support compliance test}
First, we present the results of the compliance of the torso support bar. The recorded data by the force gauge and IMU were manually aligned. The relation between the measured force and bending angle of the compliant segment is shown in Fig. \ref{fig:F-angle}. The blue line ``Body weight equivalent" represents the force needed to support 100\% of the upper body weight calculated based on anthropomorphic data for a person with body weight of 74.1 kg and body height of 174 cm, the brown line  ``Measurement" shows the measured resistive force of the torso bar during the test, and the green line ``Linear regression" shows the regression of the measured results. These results show that the compliant segment in the torso bar provides  64\% of the force for balancing the upper body weight.

As observed from Fig. \ref{fig:F-angle}, there is an effect of hysteresis. 
The effects of this hysteresis are summarized below:
1) For velocity magnitude, since it is controlled by the torso bending angle (not force), users can control it using sensory feedback of the bending angle. For velocity direction control, there is no mismatch between flexion and extension because the direction is controlled by the pressure center. Although the force decreases during the extension phase, the center remains the same.
2) The hysteresis effect represents a mismatch between simulations and real implementation, potentially affecting vibration (control stability) associated with the stiffness of the compliant segment.
3) In the extension phase, the supporting force provided by the compliant segment is lower than expected, which might be an issue for patients with weak muscles. This will be further investigated with patients having different injury levels.
Currently, users have not reported control difficulties due to hysteresis, indicating they can easily compensate for it. Since severe effects from hysteresis have not been observed, it is not compensated for in the current version of the system.

   \begin{figure}[!t]
      \centering
        \includegraphics[width=1.0\linewidth]{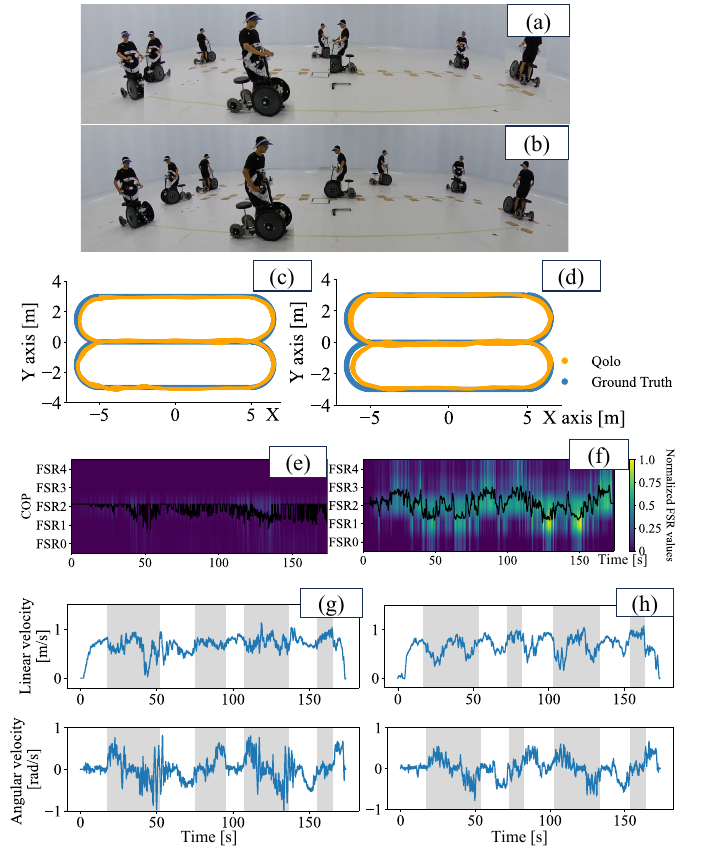}
		
      \caption{The result of one participant controlling the device navigating. (a), (c), (e), (g): joystick; (b), (d), (f), (h): torso control. (a), (b) motion sequences; (c), (d) recorded trajectory of Qolo and ground truth; (e), (f) heatmap of normalized FSR values and computed COP; (g), (h) velocity change with time, gray area indicates the path with obstacles.}
      \label{fig16} 
      \vspace{-0.4cm}
   \end{figure}

\begin{figure*}[!t]
\centering
    \includegraphics[width=0.9\linewidth]{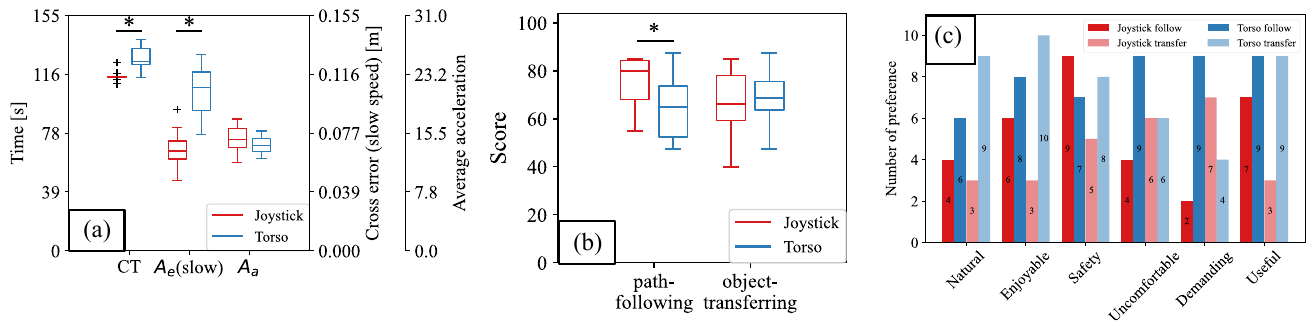}
\caption{(a) Result of $CT$, $A_a$ and $A_e$. Statistical comparisons are made between two interfaces. $* \rightarrow p<0.05$. (b) SUS questionnaire result, $* \rightarrow p<0.05$. (c) experiment-specific questionnaire result.}
\label{exp:result} 
\vspace{-0.4cm}
\end{figure*}

\vspace{-0.3cm}
\subsection{Path-following experiment}
It took 10-15 minutes on average for the participants to get used to the torso control interface. A snapshot of a participant controlling the device at a constant slow speed is shown in Fig. \ref{fig16}(a), (b). The heatmap of normalized FSR values and the computed COP are shown in Fig. \ref{fig16}(e), (f).
A trajectory and velocity example of one participant controlling the device at a constant slow speed is shown in Fig. \ref{fig16} (c), (d), (g), (h) individually.
A summary of $CT$, $A_e$ and $A_a$ is shown in Fig. \ref{exp:result} (a). Significant differences between the two interfaces were observed for $CT$ (with $p = 0.0039$) and $A_e$ (with $p = 0.0039$).
For using the torso control interface, the participants spent $10\%$ more time than using the joystick ($126.8 \pm 7.5s$ compared to $115.3 \pm 3.4s$).
The average cross error of torso control at slow speed was $0.116 \pm 0.04m$ which is significantly higher than that of the joystick ($0.067 \pm 0.01m$). However, considering the size of the device was $0.75m$, $0.116 m$, the cross error value of either interface was small relative to the size of the device.
Torso control showed slightly overall better performance ($13.98\pm1.1$) in terms of average acceleration ($4.9\%$ less) than the joystick ($14.70 \pm 1.7$), with a perceptible delay as a compromise. According to feedback from the participants, this delay while perceptible was acceptable during the control of the device.

The SUS questionnaire result is shown in Fig. \ref{exp:result} (b). There was a significant difference between the two interfaces (with $p = 0.0273$).
The usability score of torso control ($64.25\pm12.8$) was $15\%$ lower than that of the joystick ($75.75\pm9.7$). 

      \begin{figure}[!t]
      \centering
      \includegraphics[width=0.95\linewidth]{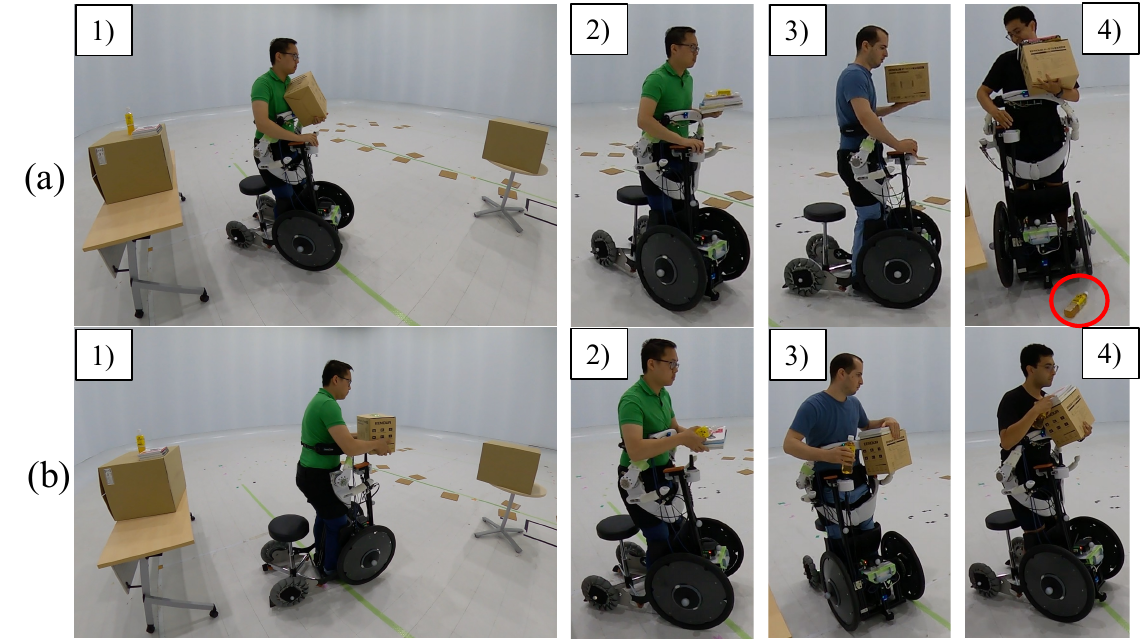}
      \caption{Object-transferring experiment, with participants holding objects. (a) The participants used joystick control. In 3), the participant had to hold the heavy object with one hand due to occupancy of the other hand, in 4), the participant dropped the water bottle (in a red circle) accidentally. (b) The participants used torso control. Participants transferred the objects with better confidence.}
      \label{exp:r-l} 
      \vspace{-0.4cm}
   \end{figure} 

\vspace{-0.3cm}
\subsection{Object-transferring experiment}
The result of the SUS questionnaire is shown in Fig. \ref{exp:result} (b). For the object-transferring task, the SUS score of the joystick decreased to $67.25\pm13.5$ compared with the path-following task and that of torso control increased to $69.5\pm11.8$, with no significant difference. 
Most participants stated that using the torso control interface makes the transferring task easier since both hands can be used. 
One female participant stated that using the joystick was easier because she struggled to balance her body while holding the 5kg object with the torso control. 

In summary, using the joystick to operate the interface is easy but holding the objects requires additional attention, while for the torso control holding the object is easier but controlling the torso for operating the interface while holding objects requires additional effort.
The object-transferring experiment demonstrates the practical value of torso control, to enable users to use their hands for additional tasks while controlling the mobility device.

\vspace{-0.3cm}
\subsection{Experiment-specific questions and interview}
Results of the device-specific questionnaire are shown in Fig. \ref{exp:result} (c). For naturalness, enjoyment, and usefulness, more participants gave a higher score to the torso control. For safety, comfort, and demand, the joystick was rated better. 
After the object-transferring experiment, for all aspects, the score of the torso control increased, while that of the joystick decreased.

For the interview, most participants liked the hands-free feature, especially in the object-transferring experiment. At the same time, they thought they could perform better with more practice using the torso control interface.
Some participants mentioned that a higher bar may be better. Especially for one female participant, who mentioned that a more adjustable bar would be easier for females.
One participant mentioned that having visual feedback would be helpful for torso control, the same as when using the joystick.
It was mentioned also by one participant that it was harder to stop precisely when using the torso control, which might be two reasons: one is the extra delay in torso control, and the second one is that reaction speed is slower when using the torso for control than when using the hand and fingers.

\vspace{-0.3cm}
\section{Overall Discussion and Conclusion}\label{sec_dis}
In this work, we use the central torso of the human as the control platform. We proposed a torso-based control interface with compliant support for assistive mobility devices. The interface is easy to wear and serves two to support the user's upper body weight and to control the device. The stiffness of the support mechanism and the control algorithm are designed as an integrated unit to enable stable and intuitive control.
The control method is based on a two-layer mapping that integrates decoupled control for velocity direction and magnitude.

The compliant segment provides approximately 64\% of the force needed to fully support the upper body weight of a 74.1kg user in the developed prototype. This can be further improved by optimizing the design for specific users. 

Experiments were conducted to verify the effectiveness of the proposed interface. The torso control interface showed comparable results with the joystick in terms of $CT$, $A_e$, $A_a$, and SUS. The various conditions in the experiment setup demonstrated the wide adaptability of the torso control, including different genders (male and female), terrains (flat and uneven), and speeds (slow and fast).
In the object-transferring experiment, participants used their arms to transfer objects and moved their heads to look at different locations, yet they were still able to control the device smoothly. This indicates that using the torso as the control input does not interfere with users' daily activities, unlike using other body parts for control input.
Due to safety concerns, we did not ask participants to try more extreme cases, such as transferring a vase of water or a large object (e.g., $50 \times 50 \times 20 \mathrm{~cm}$). However, we believe torso control would have advantages when manipulating fragile, large, or multiple objects. In the current setup, one user accidentally dropped a water bottle due to being hand-occupied with the joystick.
Participants showed a high preference for the torso control interface in experiment-specific questionnaires and interviews. The main limitation of torso control is lower position control precision compared to a joystick. It also requires more physical effort than joystick control.
Compared with most other hands-free solutions, such as BCI, sip-and-puff, or shoulder-based control, torso control offers advantages such as ease of wear, robustness, and minimal interference with the user's daily social life.

In the current experiment, all users employed a similar setup. Since the standing mobility device is single-oriented, the hardware and software can be calibrated more effectively to enhance user-friendliness. Moreover, although the learning curve was not the focus of this work, most participants adapted to the proposed interface within 15 minutes, and further improvement is expected with more practice.
As a device with human riding, several safety measures have been implemented. For example, an emergency trigger is designed to activate if the user falls, and the system will decelerate and stop if it detects a bending angle beyond 40 degrees. Healthcare professionals have noted that the torso support bar enhances safety compared to \cite{Eguchi2018} and \cite{Paez2018}. Additionally, the compliant segment of the bar makes it more comfortable and reduces pressure on the user's skin. However, compared to \cite{Paez2018}, the forward supporting force is reduced. Considering that wheelchair users, including those with SCI, often have weaker or paralyzed lower trunk muscles, an additional system would be necessary for safety. In case they cannot remain upright without any support, it will be difficult to apply the proposed method. In an emergency, the torso control should be turned off electronically to maintain the supporting function of the torso bar as much as possible.

In this paper, we focused the test on unimpaired users. For persons with SCI, the level of injury can significantly affect their ability to control the hip joint muscles \cite{zhang2022design}. Therefore, the target of this work was set to below T12. In the future, we plan to conduct different experiments with patients having different levels of SCI injury to assess their ability to control this system based on their specific injury levels.

\vspace{-0.3cm}

% use section* for acknowledgment
\section*{Acknowledgment}
The authors would like to thank Dr. Hideki Kadone and Dr. Yukiyo Shimizu for their invaluable contributions to this project from its inception. Their extensive end-user studies on torso control, along with their expert advice and assessments as healthcare professionals, were instrumental in shaping the design of the interface. Additionally, the authors wish to thank Dr. Luis Ccorimanya and Xiaoxi Zhang for their assistance with the experimental setup, and Dr. Mehdi Benallegue for his advice on system dynamics analysis. 
This work was supported by the Ministry of Education, Culture, Sports, Science and Technology of Japan under grant numbers 22J13072, 22KJ0415, and 23H00485.

\vspace{-0.3cm}
\appendix \label{Appendix}
 The kinematics of the system and its derivative are:

\begin{equation}
\begin{aligned}
& x_m=x-l \sin \theta, && y_m=l \cos \theta \\
& \dot{x}_m=\dot{x}-l \cos \theta \cdot \dot{\theta}, && \dot{y}_m=-l \dot{\theta} \sin \theta
\end{aligned}
\end{equation}

The potential energy is:
\begin{equation}
V=m g y_m+\frac{1}{2} k(l \sin \theta)^2=m g l \cos \theta+\frac{1}{2} k l^2 \sin ^2 \theta
\end{equation}

The kinetic energy is:
\begin{equation}
    T =\frac{1}{2}(m+M) \dot{x}^2+\frac{1}{2} m l^2 \dot{\theta}^2-m l \dot{\theta} \dot{x} \cos \theta,
\end{equation}

according to Lagrangian formulation:
\begin{equation}
    \begin{aligned}
    \begin{array}{l}
    L = T-V, \:
    \frac{d}{d t}\left(\frac{\partial L}{\partial \dot{q}_i}\right)-\frac{\partial L}{\partial q_i}=Q_i
    \end{array}
    \end{aligned},
\end{equation}

we can obtain the dynamics equations in (\ref{equ:dynamics}):
\begin{equation}
\vspace{-0.2cm}
    \begin{array}{l}
    \begin{aligned}
\ddot{x} &= \frac{m g \sin \theta \cos \theta - k_s l \sin \theta \cos^2 \theta}{M + m \sin^2 \theta} \\
&\quad - \frac{m l \dot{\theta}^2 \sin \theta + k_c l \cos^2 \theta \cdot \dot{\theta}}{M + m \sin^2 \theta}  + \frac{h \cos \theta + u - k_d \dot{x}}{M + m \sin^2 \theta} \\
\ddot{\theta} &= \frac{\ddot{x} \cos \theta + g \sin \theta}{l} \\
&\quad - \frac{k_s \sin \theta \cos \theta + k_c \cos \theta \cdot \dot{\theta}}{m}  + \frac{h}{ml}
\end{aligned}
    \end{array}
     \label{equ:dynamics}.
\end{equation}

% Can use something like this to put references on a page
% by themselves when using endfloat and the captionsoff option.
\ifCLASSOPTIONcaptionsoff
  \newpage
\fi

\bibliographystyle{IEEEtran}
\bibliography{IEEEabrv,Qolo_torso}

% Generated by IEEEtran.bst, version: 1.14 (2015/08/26)
\begin{thebibliography}{10}
\providecommand{\url}[1]{#1}
\csname url@samestyle\endcsname
\providecommand{\newblock}{\relax}
\providecommand{\bibinfo}[2]{#2}
\providecommand{\BIBentrySTDinterwordspacing}{\spaceskip=0pt\relax}
\providecommand{\BIBentryALTinterwordstretchfactor}{4}
\providecommand{\BIBentryALTinterwordspacing}{\spaceskip=\fontdimen2\font plus
\BIBentryALTinterwordstretchfactor\fontdimen3\font minus \fontdimen4\font\relax}
\providecommand{\BIBforeignlanguage}[2]{{%
\expandafter\ifx\csname l@#1\endcsname\relax
\typeout{** WARNING: IEEEtran.bst: No hyphenation pattern has been}%
\typeout{** loaded for the language `#1'. Using the pattern for}%
\typeout{** the default language instead.}%
\else
\language=\csname l@#1\endcsname
\fi
#2}}
\providecommand{\BIBdecl}{\relax}
\BIBdecl

\bibitem{whill}
{WHILL}, ``Whill,'' \url{https://whill.inc/jp}, 2023, [Online; accessed 20-February-2023].

\bibitem{LEVO}
{LEVO AG}, ``Levo history,'' \url{https://levousa.com/history/}, 2020, [Online; accessed 2020-01-31].

\bibitem{gyrolift}
{GYROLIFT}, ``Gyrolift,'' 2017, \url{http://www.gyrolift.fr/}, Last accessed on 2020-01-31.

\bibitem{niijima2019real}
S.~Niijima, Y.~Sasaki, and H.~Mizoguchi, ``Real-time autonomous navigation of an electric wheelchair in large-scale urban area with 3d map,'' \emph{Advanced Robotics}, vol.~33, no.~19, pp. 1006--1018, 2019.

\bibitem{Chen_2021}
Y.~Chen, D.~F. Paez-Granados, B.~Leme, and K.~Suzuki, ``\BIBforeignlanguage{en}{Virtual landmark-based control of docking support for assistive mobility devices},'' \emph{\BIBforeignlanguage{en}{IEEE/ASME Transactions on Mechatronics}}, vol.~26, no.~4, p. 2007–2015, Aug 2021.

\bibitem{ashok2017high}
S.~Ashok, ``High-level hands-free control of wheelchair--a review,'' \emph{Journal of medical engineering \& technology}, vol.~41, no.~1, pp. 46--64, 2017.

\bibitem{deng2019bayesian}
X.~Deng, Z.~L. Yu, C.~Lin, Z.~Gu, and Y.~Li, ``A bayesian shared control approach for wheelchair robot with brain machine interface,'' \emph{IEEE Transactions on Neural Systems and Rehabilitation Engineering}, vol.~28, no.~1, pp. 328--338, 2019.

\bibitem{chenTorso}
Y.~Chen, D.~Paez-Granados, H.~Kadone, and K.~Suzuki, ``Control interface for hands-free navigation of standing mobility vehicles based on upper-body natural movements,'' in \emph{2020 IEEE/RSJ International Conference on Intelligent Robots and Systems (IROS)}, 2020, pp. 11\,322--11\,329.

\bibitem{ji2021design}
J.~Ji, W.~Chen, W.~Wang, and J.~Xi, ``Design and control of an omni-directional robotic walker based on human--machine interaction,'' \emph{IEEE Access}, vol.~9, pp. 111\,358--111\,367, 2021.

\bibitem{zhao2020smart}
X.~Zhao, Z.~Zhu, M.~Liu, C.~Zhao, Y.~Zhao, J.~Pan, Z.~Wang, and C.~Wu, ``A smart robotic walker with intelligent close-proximity interaction capabilities for elderly mobility safety,'' \emph{Frontiers in neurorobotics}, vol.~14, p. 575889, 2020.

\bibitem{dahmani2020intelligent}
M.~Dahmani, M.~E. Chowdhury, A.~Khandakar, T.~Rahman, K.~Al-Jayyousi, A.~Hefny, and S.~Kiranyaz, ``An intelligent and low-cost eye-tracking system for motorized wheelchair control,'' \emph{Sensors}, vol.~20, no.~14, p. 3936, 2020.

\bibitem{rechy2012head}
E.~J. Rechy-Ramirez, H.~Hu, and K.~McDonald-Maier, ``Head movements based control of an intelligent wheelchair in an indoor environment,'' in \emph{2012 IEEE International Conference on Robotics and Biomimetics (ROBIO)}.\hskip 1em plus 0.5em minus 0.4em\relax IEEE, 2012, pp. 1464--1469.

\bibitem{grewal2018sip}
H.~S. Grewal, A.~Matthews, R.~Tea, V.~Contractor, and K.~George, ``Sip-and-puff autonomous wheelchair for individuals with severe disabilities,'' in \emph{2018 9th IEEE Annual Ubiquitous Computing, Electronics \& Mobile Communication Conference (UEMCON)}.\hskip 1em plus 0.5em minus 0.4em\relax IEEE, 2018, pp. 705--710.

\bibitem{Simpson2008}
T.~Simpson~$^*$, C.~Broughton, M.~J.~A. Gauthier, and A.~Prochazka, ``Tooth-click control of a hands-free computer interface,'' \emph{IEEE Transactions on Biomedical Engineering}, vol.~55, no.~8, pp. 2050--2056, 2008.

\bibitem{kong2019stand}
F.~Kong, M.~N. Sahadat, M.~Ghovanloo, and G.~D. Durgin, ``A stand-alone intraoral tongue-controlled computer interface for people with tetraplegia,'' \emph{IEEE transactions on biomedical circuits and systems}, vol.~13, no.~5, pp. 848--857, 2019.

\bibitem{lontis2021wheelchair}
E.~R. Lontis, B.~Bentsen, M.~Gaihede, F.~Biering-S{\o}rensen, and L.~N.~A. Struijk, ``Wheelchair control with inductive intra-oral tongue interface for individuals with tetraplegia,'' \emph{IEEE Sensors Journal}, vol.~21, no.~20, pp. 22\,878--22\,890, 2021.

\bibitem{plotkin2010sniffing}
A.~Plotkin, L.~Sela, A.~Weissbrod, R.~Kahana, L.~Haviv, Y.~Yeshurun, N.~Soroker, and N.~Sobel, ``Sniffing enables communication and environmental control for the severely disabled,'' \emph{Proceedings of the National Academy of Sciences}, vol. 107, no.~32, pp. 14\,413--14\,418, 2010.

\bibitem{nishimori2007voice}
M.~Nishimori, T.~Saitoh, and R.~Konishi, ``Voice controlled intelligent wheelchair,'' in \emph{SICE Annual Conference 2007}.\hskip 1em plus 0.5em minus 0.4em\relax IEEE, 2007, pp. 336--340.

\bibitem{choi2006new}
K.~Choi, M.~Sato, and Y.~Koike, ``A new, human-centered wheelchair system controlled by the emg signal,'' in \emph{The 2006 IEEE International Joint Conference on Neural Network Proceedings}.\hskip 1em plus 0.5em minus 0.4em\relax IEEE, 2006, pp. 4664--4671.

\bibitem{Chronus}
{Chronus Robotics}, ``Chronus kim1,'' \url{https://chronusrobotics.com/}, 2023, [Online; accessed 1st-July-2023].

\bibitem{UNI-ONE}
{Honda Motor Co., Ltd.}, ``Uni-one,'' \url{https://www.honda.co.jp/future/UNI-ONE/}, 2023, [Online; accessed 1st-July-2023].

\bibitem{Nguyen2004}
H.~G. Nguyen, J.~Morrell, K.~D. Mullens, A.~B. Burmeister, S.~Miles, N.~Farrington, K.~M. Thomas, and D.~W. Gage, ``{Segway robotic mobility platform},'' \emph{Mobile Robots XVII}, vol. 5609, no. February 2002, p. 207, 2004.

\bibitem{wakita2012riding}
Y.~Wakita, Y.~Kato, M.~Sakakibara, and K.~Inoue, ``Riding type vehicle and method of controlling riding type vehicle,'' Dec.~25 2012, uS Patent 8,340,869.

\bibitem{Thorp2016}
E.~B. Thorp, F.~Abdollahi, D.~Chen, A.~Farshchiansadegh, M.-H. Lee, J.~Pedersen, C.~Pierella, E.~J. Roth, S.~G. Ismael, and F.~A. Mussa-Ivaldi, ``{Upper Body-Based Power Wheelchair Control Interface for Individuals with Tetraplegia},'' \emph{IEEE Transactions on Neural Systems and Rehabilitation Engineering}, vol.~24, no.~2, pp. 249--260, 2016.

\bibitem{moore2018clinically}
K.~L. Moore and A.~F. Dalley, \emph{Clinically oriented anatomy}.\hskip 1em plus 0.5em minus 0.4em\relax Wolters kluwer india Pvt Ltd, 2018.

\bibitem{cromwell2001sagittal}
R.~L. Cromwell, T.~K. Aadland-Monahan, A.~T. Nelson, S.~M. Stern-Sylvestre, and B.~Seder, ``Sagittal plane analysis of head, neck, and trunk kinematics and electromyographic activity during locomotion,'' \emph{Journal of Orthopaedic \& Sports Physical Therapy}, vol.~31, no.~5, pp. 255--262, 2001.

\bibitem{Eguchi2018}
Y.~Eguchi, H.~Kadone, and K.~Suzuki, ``Standing mobility device with passive lower limb exoskeleton for upright locomotion,'' \emph{IEEE/ASME Transactions on Mechatronics}, vol.~23, no.~4, pp. 1608--1618, 2018.

\bibitem{Paez2018}
D.~Paez-Granados, H.~Kadone, and K.~Suzuki, ``{Unpowered Lower-Body Exoskeleton with Torso Lifting Mechanism for Supporting Sit-to-Stand Transitions},'' in \emph{IEEE International Conference on Intelligent Robots and Systems}.\hskip 1em plus 0.5em minus 0.4em\relax Madrid: IEEE Xplorer, 2018, pp. 2755--2761.

\bibitem{chen2023wemo}
Y.~Chen, T.~Kuwahara, Y.~Nishimura, and K.~Suzuki, ``Wemo: A prototype of a wearable mobility device adapting to user’s natural posture changes,'' \emph{Sensors}, vol.~23, no.~18, p. 7683, 2023.

\bibitem{de1996adjustments}
P.~De~Leva, ``Adjustments to zatsiorsky-seluyanov's segment inertia parameters,'' \emph{Journal of biomechanics}, vol.~29, no.~9, pp. 1223--1230, 1996.

\bibitem{courtine2003human}
G.~Courtine and M.~Schieppati, ``Human walking along a curved path. i. body trajectory, segment orientation and the effect of vision,'' \emph{European Journal of Neuroscience}, vol.~18, no.~1, pp. 177--190, 2003.

\bibitem{eguchi_thesis}
\BIBentryALTinterwordspacing
Y.~Eguchi, ``A study on passive exoskeleton assisting postural transformation, and mobility device for upright locomotion,'' Ph.D. dissertation, University of Tsukuba, 2019. [Online]. Available: \url{https://cir.nii.ac.jp/crid/1110575131568029568}
\BIBentrySTDinterwordspacing

\bibitem{lewis1995ibm}
J.~R. Lewis, ``Ibm computer usability satisfaction questionnaires: psychometric evaluation and instructions for use,'' \emph{International Journal of Human-Computer Interaction}, vol.~7, no.~1, pp. 57--78, 1995.

\bibitem{zhang2022design}
T.~Zhang, C.~Ning, Y.~Li, and M.~Wang, ``Design and validation of a lightweight hip exoskeleton driven by series elastic actuator with two-motor variable speed transmission,'' \emph{IEEE Transactions on Neural Systems and Rehabilitation Engineering}, vol.~30, pp. 2456--2466, 2022.

\end{thebibliography}

  \begin{IEEEbiography}[{\includegraphics[width=1in,height=1.2in,clip,keepaspectratio]{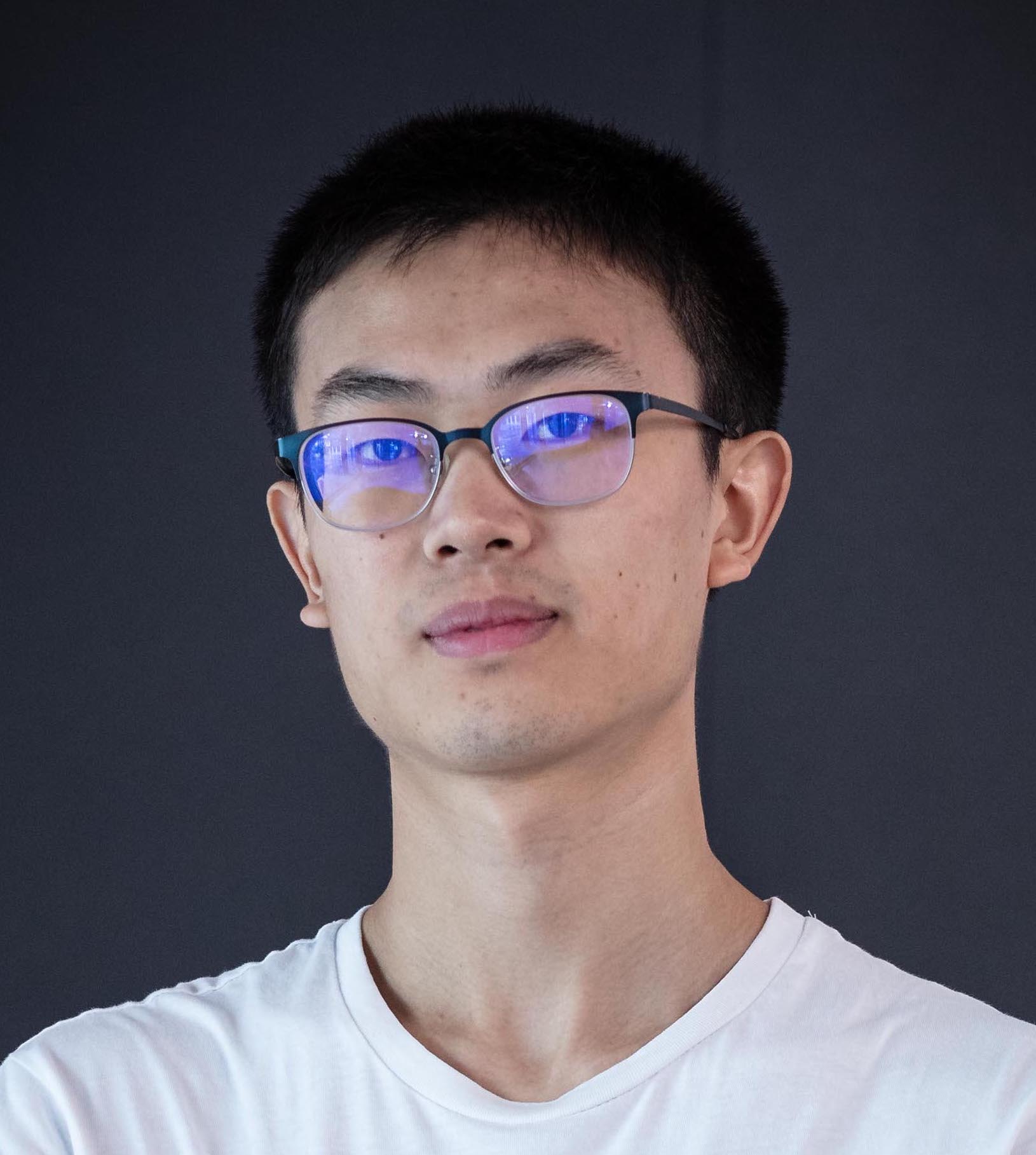}}]{Yang Chen}
received the B.Eng. in mechanical engineering from Jilin University, China, in 2017 and his master and Ph. D. degree in human informatics from the University of Tsukuba, Japan, in 2020 and 2023 respectively. He is currently a postdoctoral researcher with the University of Tsukuba. His research interests include human augmentation, mobile robotics and human-robot interaction. 
  \end{IEEEbiography}
  
\begin{IEEEbiography}[{\includegraphics[width=1in,height=1.2in,clip,keepaspectratio]{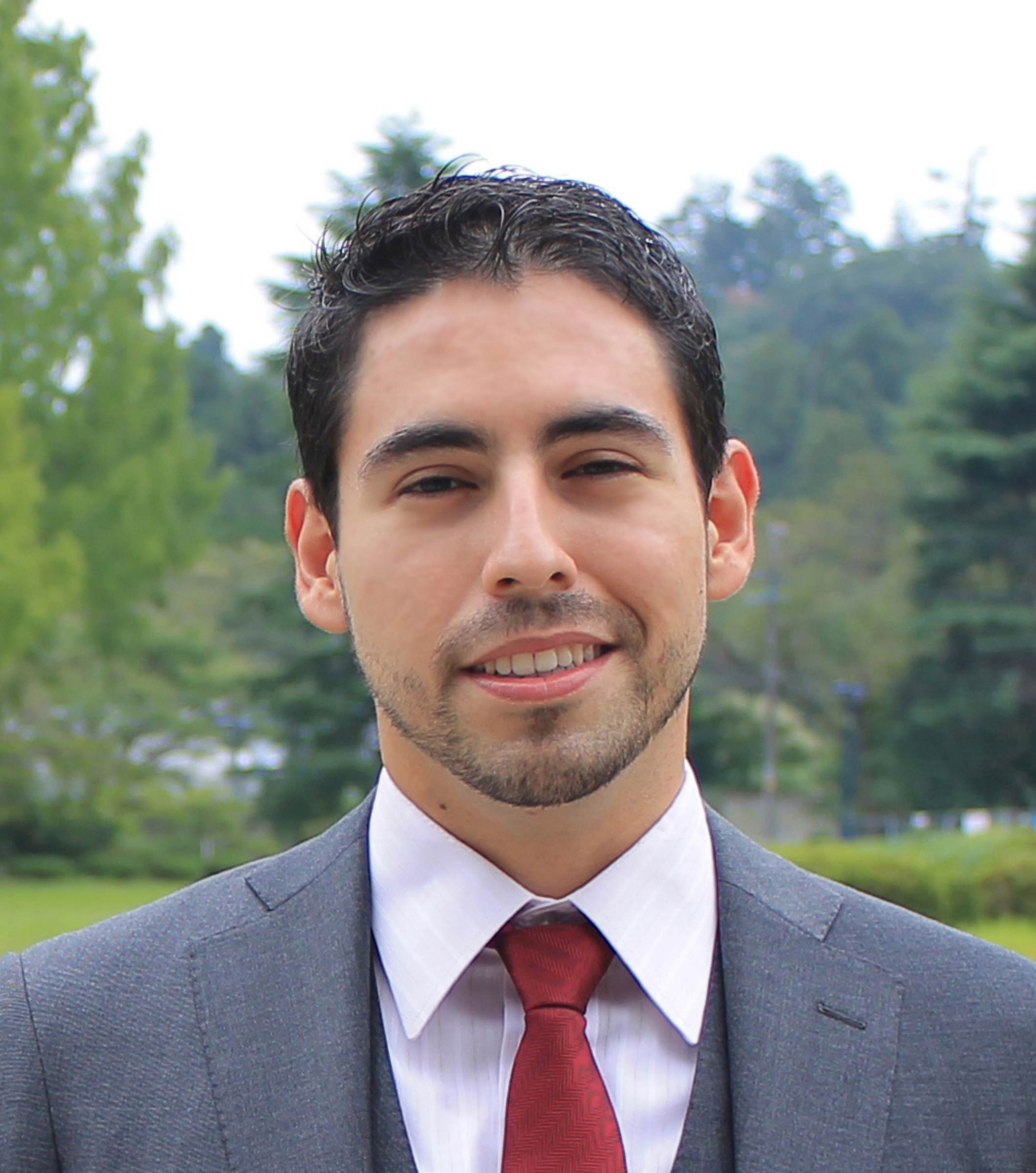}}]{Diego F. Paez-Granados}
received the Ph.D. degree in bioengineering and robotics from Tohoku University in 2017. He is currently a Researcher with the Learning Algorithms and Systems Laboratory, Swiss Federal Institute of Technology Lausanne, Lausanne, Switzerland, and a Visiting Researcher with the University of Tsukuba, Tsukuba, Japan. His research interests include physical and cognitive human modeling, compliance in humanrobot interaction, shared control for robot navigation, and soft-robot design and control with human in the loop.
\end{IEEEbiography}

\begin{IEEEbiography}[{\includegraphics[width=1in,height=1.2in,clip,keepaspectratio]{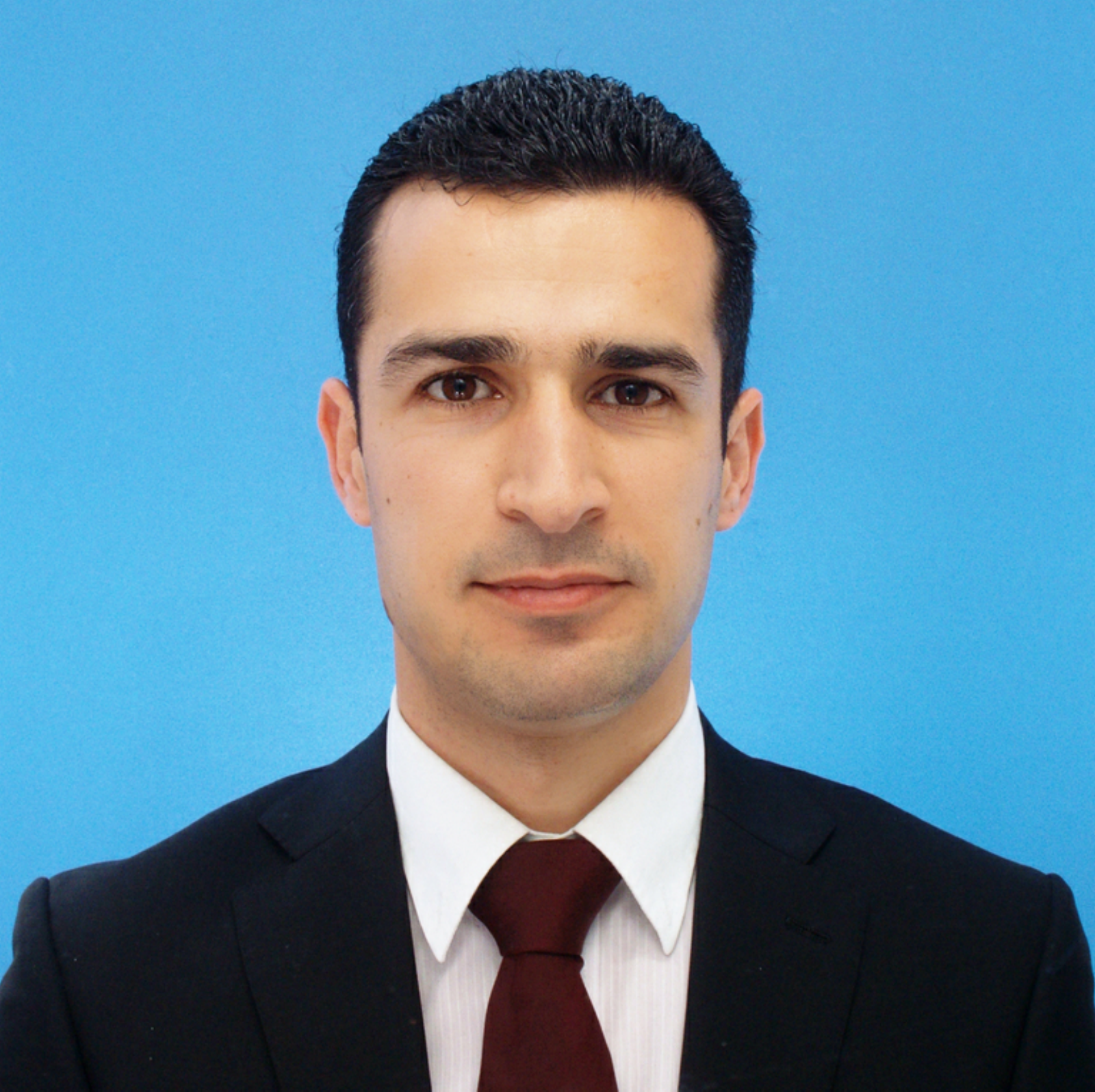}}]{Modar Hassan} 
received the M.E. and Ph.D. degrees in engineering and the M.S. degree in medical sciences from the University of Tsukuba, Tsukuba, Japan, in 2013, 2016, and 2016, respectively. 
% From 2010 to 2011, he was a Lecturer in mechatronics with Tishreen University. 
From 2016 to 2021, he was a Postdoctoral Researcher with the University of Tsukuba, where he is currently an Assistant Professor with the Faculty of Engineering, Information and Systems. His research interests include assistive robotics, wearable devices, esports and para-esports, orthotics, prosthetics, biomechanics, and motor control.
\end{IEEEbiography}

\begin{IEEEbiography}[{\includegraphics[width=1in,height=1.2in,clip,keepaspectratio]{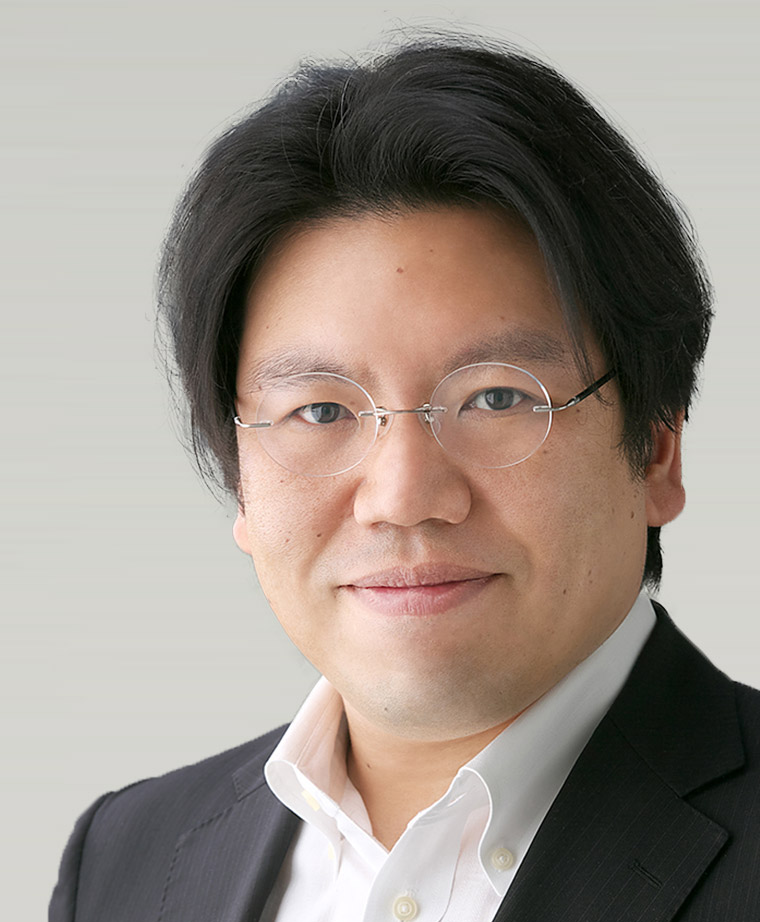}}]{Kenji Suzuki}
received the Ph.D. degree in pure and applied physics from Waseda University, Tokyo, Japan, in 2003. He is currently a Full Professor with the Center for Cybernics Research and the Principal Investigator with Artificial Intelligence Laboratory, University of Tsukuba, Tsukuba, Japan. His research interests include wearable robotics and devices, affective computing, social robotics, and assistive robotics. Dr. Suzuki has been an elected AdCom Member of IEEE Robotics and Automation Society since 2019.
\end{IEEEbiography}

\end{document}